\documentclass[sigconf, nonacm]{acmart}

\usepackage{tabularx}
\usepackage{graphicx}
\usepackage{hyperref}
\usepackage{balance}  

\usepackage{amsmath}

\usepackage{color}
\usepackage{multirow}
\usepackage[export]{adjustbox}
\usepackage{tikz}
\usetikzlibrary{matrix, positioning}
\usepackage{bbm}
\usepackage{array}
\usepackage{caption}
\usepackage{subcaption}
\usepackage{enumitem}

\newcommand{\stitle}[1]{\noindent\textbf{#1}}

\usepackage{algorithmicx}
\usepackage{algorithm}
\usepackage{algpseudocode}

\newcommand\vldbdoi{10.14778/3681954.3681993}
\newcommand\vldbpages{3192 - 3200}
\newcommand\vldbvolume{17}
\newcommand\vldbissue{11}
\newcommand\vldbyear{2024}
\newcommand\vldbauthors{\authors}
\newcommand\vldbtitle{\shorttitle} 
\newcommand\vldbavailabilityurl{https://github.com/Carrieww/GraphHT}
\newcommand\vldbpagestyle{empty} 

\definecolor{myGreen}{RGB}{0, 161, 8}

\newtheorem{definition}{Definition}
\newtheorem{theorem}{Theorem}

\newcommand{\hide}[1]{}

\begin{document}

\title{A Sampling-based Framework for Hypothesis Testing on Large Attributed Graphs}

\author{Yun Wang}
\affiliation{%
  \institution{The
     University of Hong Kong}
}
\email{carrie07@connect.hku.hk}

\author{Chrysanthi Kosyfaki}
\affiliation{%
  \institution{The
     University of Hong Kong}
}
\email{kosyfaki@cs.hku.hk}

\author{Sihem Amer-Yahia}
\affiliation{%
  \institution{CNRS, Univ. Grenoble Aples}
}
\email{sihem.amer-yahia@univ-grenoble-alpes.fr}

\author{Reynold Cheng}
\affiliation{%
  \institution{The
     University of Hong Kong}
}
\email{ckcheng@cs.hku.hk}


\begin{abstract}
Hypothesis testing is a statistical method used to draw conclusions about populations from sample data, typically represented in tables. With the prevalence of graph representations in real-life applications, hypothesis testing on graphs is gaining importance. In this work, we formalize node, edge, and path hypotheses on attributed graphs. We develop a sampling-based hypothesis testing framework, which can accommodate existing hypothesis-agnostic graph sampling methods. To achieve accurate and time-efficient sampling, we then propose a Path-Hypothesis-Aware SamplEr, PHASE, an $m$-dimensional random walk that accounts for the paths specified in the hypothesis. We further optimize its time efficiency and propose $\text{PHASE}_{\text{opt}}$. Experiments on three real datasets demonstrate the ability of our framework to leverage common graph sampling methods for hypothesis testing, and the superiority of hypothesis-aware sampling methods in terms of accuracy and time efficiency. 
\end{abstract}

\maketitle

\pagestyle{\vldbpagestyle}
\begingroup\small\noindent\raggedright\textbf{PVLDB Reference Format:}\\
\vldbauthors. \vldbtitle. PVLDB, \vldbvolume(\vldbissue): \vldbpages, \vldbyear.\\
\href{https://doi.org/\vldbdoi}{doi:\vldbdoi}
\endgroup

\begingroup
\renewcommand\thefootnote{}\footnote{\noindent
This work is licensed under the Creative Commons BY-NC-ND 4.0 International License. Visit \url{https://creativecommons.org/licenses/by-nc-nd/4.0/} to view a copy of this license. For any use beyond those covered by this license, obtain permission by emailing \href{mailto:info@vldb.org}{info@vldb.org}. Copyright is held by the owner/author(s). Publication rights licensed to the VLDB Endowment. \\
\raggedright Proceedings of the VLDB Endowment, Vol. \vldbvolume, No. \vldbissue\ %
ISSN 2150-8097. \\
\href{https://doi.org/\vldbdoi}{doi:\vldbdoi} \\
}\addtocounter{footnote}{-1}\endgroup

\ifdefempty{\vldbavailabilityurl}{}{
\vspace{.3cm}
\begingroup\small\noindent\raggedright\textbf{PVLDB Artifact Availability:}\\
The source code, data, and/or other artifacts have been made available at \url{\vldbavailabilityurl}.
\endgroup
}

\section{Introduction}\label{sec:intro}

Hypothesis testing finds widespread application in various domains such as marketing, healthcare, and social science~\cite{newey1994large}. Hypotheses are usually tested on representative samples since it is impractical, or even impossible to extract data from entire populations due to size, accessibility or cost. For instance, snowball sampling has been proven effective in accessing a hidden population like drug abusers through a chain-referral mechanism ~\cite{waters2015snowball, goodman1961snowball}. In this paper, we study hypothesis testing on graphs. 

\begin{figure}[!t]
    \centering
    \includegraphics[width=\columnwidth]{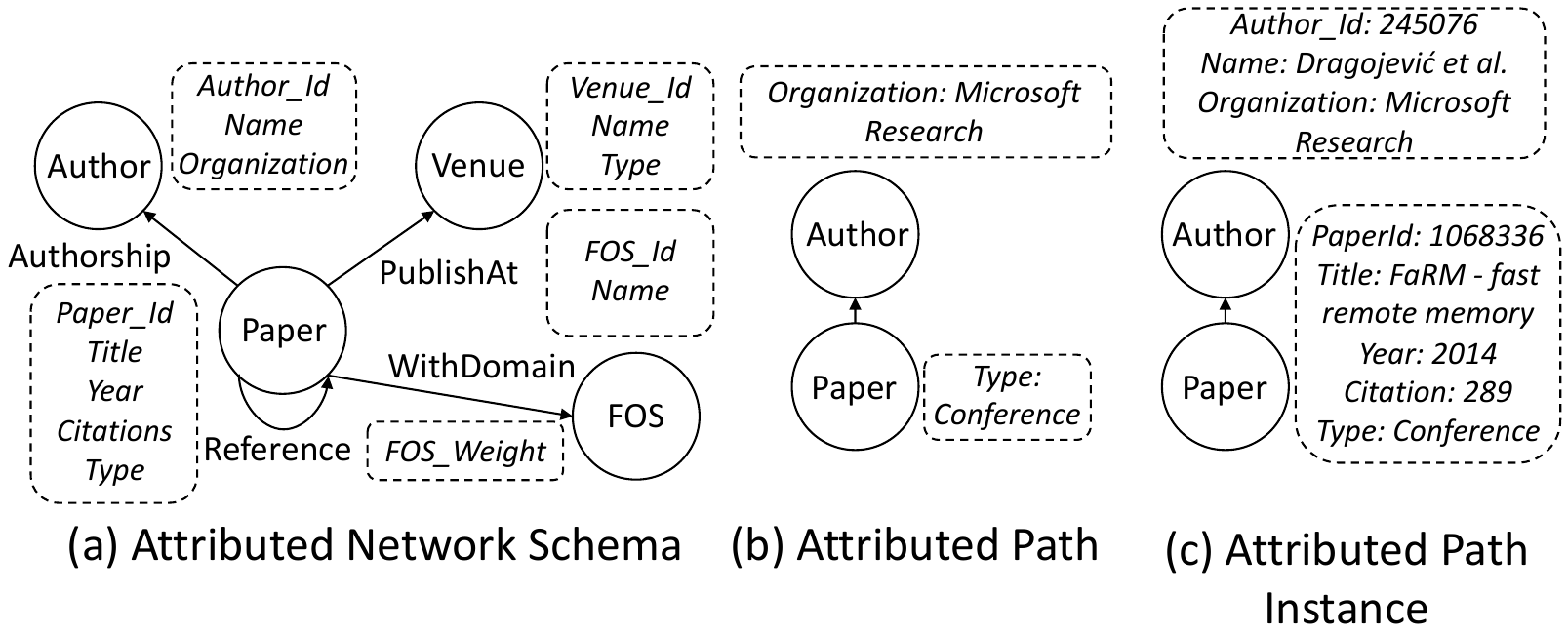}
    \caption{DBLP network schema and paths}
    \label{fig:network_schema}
\end{figure}

Graphs can represent real-world applications such as social, bibliographic, and transportation networks where entities are nodes and edges reflect relationships between them. We focus on attributed graphs that contain multiple types of nodes and edges, each labeled with attributes that provide valuable information for graph analytics~\cite{croft2011hypothesis, DBLP:journals/vldb/Omidvar-Tehrani19}. 
In a bibliographic network like DBLP (see Figure~\ref{fig:network_schema}a), hypotheses expressed on nodes, edges, and paths unveil interesting aspects of collaboration behavior and research trends based on citation patterns. 
A node hypothesis examines attributes of a single node type. For example, ``the average citation of conference papers is greater than average'' concerns the citation attribute of the conference paper nodes.
Path hypotheses consider attributes along a defined path, such as a path connecting authors from Microsoft to their conference papers (see Figure \ref{fig:network_schema}b and \ref{fig:network_schema}c for the path and an path instance, respectively). An example of path hypothesis is ``the minimum number of citations of conference papers written by Microsoft researchers is greater than average''.

\stitle{Objectives and Challenges. } 
Given an attributed graph $\mathcal{G}$, our aim is to verify a hypothesis $H$ and return a true or false result. We aim to achieve two objectives: {\bf O1} - ensure hypothesis testing is as accurate as possible; {\bf O2} - minimize execution time.
To the best of our knowledge, there is no existing literature addressing hypotheses on attributed graphs. 
A straightforward approach is to sample a subgraph \(\mathcal{S}\) from $\mathcal{G}$ using {\em hypothesis-agnostic} sampling methods~\cite{DBLP:conf/www/MaiyaB10,DBLP:conf/icde/LiWL0LXL19,DBLP:journals/corr/cs-NI-0103016}. Sampling is important when the full graph is not accessible or when the hypothesis targets very specific nodes, edges, or paths that are time-consuming to collect. However, hypothesis-agnostic samplers struggle to achieve both objectives. First, hypothesis-agnostic samplers may miss {\em relevant} nodes, edges, or paths, i.e., those requested by the hypothesis, especially when the sampling budget $B$, i.e., the size of $\mathcal{S}$, is small. The irrelevant nodes, edges, or paths, which must be filtered out for hypothesis testing, can compromise accuracy. This raises the accuracy challenge behind {\bf O1}. We show in Section \ref{frm:convergence} that the {\bf hypothesis estimator}, which computes aggregate values of nodes, edges, or paths specified in $H$, {\em converges to} the ground truth in $\mathcal{G}$ as $B$ increases, resulting in accuracy approaching one. However, the rate of convergence depends on the sampler, presenting a time efficiency challenge behind {\bf O2}. By optimizing the sampler design, we aim to balance accuracy and time efficiency, enabling earlier sampling halts with accurate hypothesis testing results.

\stitle{Contributions. }
We classify hypotheses on attributed graphs into three types: node, edge, and path ones. We develop a sampling-based hypothesis testing framework, which accommodates common hypothesis-agnostic samplers, such as random node sampler~\cite{stumpf2005subnets}, random walk~\cite{DBLP:conf/infocom/GjokaKBM10}, and non-backtracking random walk~\cite{DBLP:conf/sigmetrics/LeeXE12}. 

The lack of awareness of the input hypothesis in existing sampling methods may slow down the rate of convergence of accuracy. 
Therefore, to address objectives {\bf O1} and {\bf O2}, we design PHASE, a Path-Hypothesis-Aware SamplEr. PHASE {\em is aware of} $H$ and preserves the corresponding nodes, edges, or paths. Consequently, the resulting $\mathcal{S}$ is more likely to accurately test the hypothesis. 
PHASE employs $m \geq 1$ dependent random walks with two weight functions to ensure path-hypothesis-awareness. One function prioritizes the seed selection toward the first node in the path hypothesis. The other steers the transition probability towards the nodes specified in the hypothesis. We further propose $\text{PHASE}_{\text{opt}}$ to improve the time efficiency of PHASE. $\text{PHASE}_{\text{opt}}$ adopts a non-backtracking approach to avoid selecting previously visited nodes. It also reduces computation overhead by fixing the number of neighbors to examine. We show both theoretically and empirically that, for all samplers, the hypothesis estimator {\em converges to} the ground truth as \(B\) increases, and our proposed samplers ensures earlier and smoother convergence of the hypothesis estimator.

We conduct extensive experiments on three real-world datasets: MovieLens, DBLP, and Yelp. We aim to demonstrate the effectiveness of two optimizations in $\text{PHASE}_{\text{opt}}$ compared to PHASE, and to compare the test significance, accuracy, and execution time of hypothesis-agnostic and hypothesis-aware samplers. 

We observe $\text{PHASE}_{\text{opt}}$ is at least 43 times faster than PHASE with less than 4\% accuracy difference in DBLP. For significance, $\text{PHASE}_{\text{opt}}$ consistently delivers the most precise and reliable estimates. Compared to 11 state-of-the-art hypothesis-agnostic samplers, we find $\text{PHASE}_{\text{opt}}$ excels in accuracy when $B$ is fixed across various hypothesis types and datasets. It also demonstrates robust accuracy performance, especially for difficult hypotheses, i.e., those with longer paths or fewer relevant nodes, edges, or paths in $\mathcal{G}$. Moreover, the high accuracy achieved in the shortest amount of time makes our proposed sampler usable in practice.
\section{Definitions}\label{sec:def}

\subsection{Attributed Graphs}
\begin{definition}[Attributed Graph]
    An attributed graph 
    is a directed graph $\mathcal{G} = (\mathcal{V, E})$, where $\mathcal{V}$ denotes the set of nodes and $\mathcal{E} \subseteq \mathcal{V} \times \mathcal{V}$ is the set of edges, represented by ordered pairs of nodes. It has a node type mapping function $\phi: \mathcal{V}\rightarrow \mathcal{T}$ and an edge type mapping function $\psi: \mathcal{E}\rightarrow \mathcal{R}$. 
    Each node and edge has attributes.
\end{definition}

Each node $v$ in $\mathcal{G}$ belongs to a specific node type $\phi(v) \in \mathcal{T}$, and each edge $e=(u,v)$ connecting nodes of type $\phi(u)$ to $\phi(v)$ belongs to a specific edge type $r=\psi(e) \in \mathcal{R}$. Once the relation $r$ exists, its inverse relation $r^{-1}$ naturally holds from $\phi(v)$ to $\phi(u)$. In this work, we assume each node in $\mathcal{G}$ has at least one incoming or outgoing edge. It implies the connectedness of the graph.

The DBLP Network in Figure~\ref{fig:network_schema}a is an attributed graph containing four types of nodes and four types of edges. For example, paper nodes belong to the node type $\mathtt{paper} \in \mathcal{T}$ with five attributes, including title and year. Edges from paper to field of study (FOS) nodes belong to the edge type $\mathtt {WithDomain} \in \mathcal{R}$ with a weight attribute indicating the paper's relevance to an FOS. The reverse relationship $\mathtt{WithDomain}^{-1} \in \mathcal{R}$ holds accordingly from FOS to paper nodes.

\begin{definition}[Path]
A path $\mathcal{P}$ is defined as 
\[t_1 \xrightarrow{r_1} \dots \xrightarrow{r_l} t_{l+1}\]
where the node type $t_i$ and edge type $r_i$ can repeat; $l\geq 0$ is the length of $\mathcal{P}$. When $l = 0$, $\mathcal{P}$ is a node and when $l = 1$, it is an edge.

\end{definition}

An \textbf{attributed path} is a path where each node has some attributes, referred to as \textbf{modifiers}. Figure~\ref{fig:network_schema}b presents an example of a length-one attributed path, ``conference papers written by Microsoft researchers'', and Figure~\ref{fig:network_schema}c is an instance of that path.

\begin{figure*}[t!]
    \centering
    \includegraphics[width=0.85\textwidth]{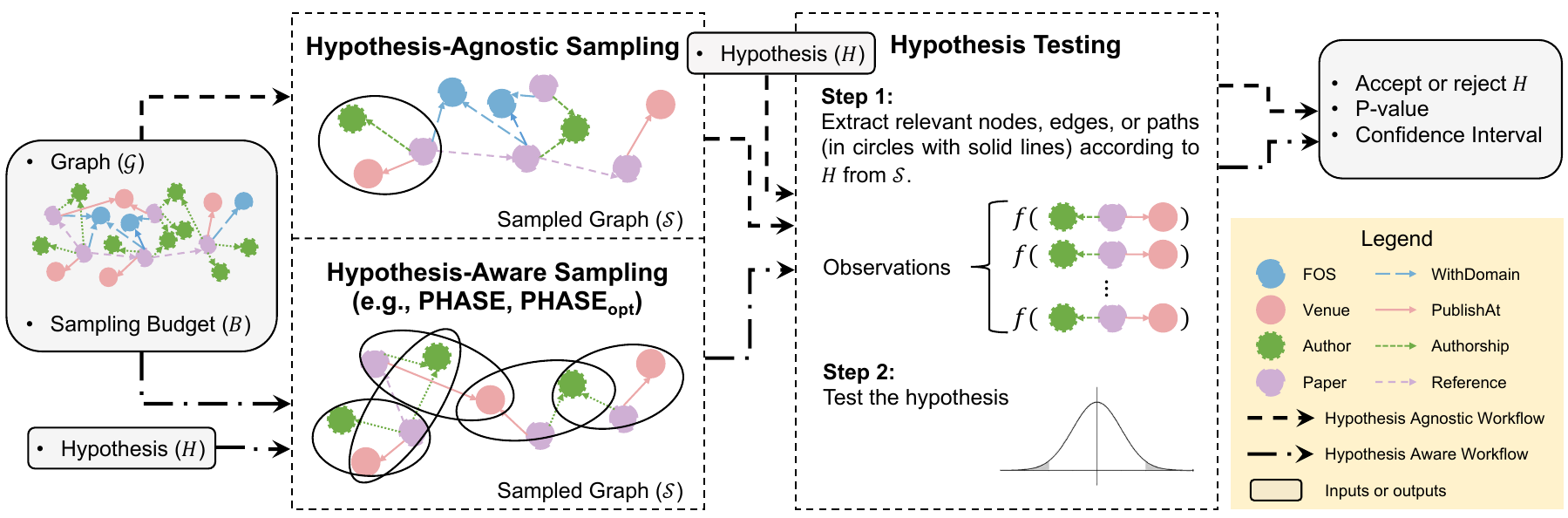}
    \caption{The Sampling-based Hypothesis Testing Framework on attributed graphs.}
    \label{fig:framework}
\end{figure*}

\subsection{Hypotheses on Attributed Graphs}
We formally define path hypotheses on attributed graphs. Node and edge hypotheses are two special cases of path hypotheses.

\begin{definition}[Path Hypothesis] 
    Given a path $\mathcal{P} = t_1 \xrightarrow{r_1} t_2 \xrightarrow{r_2} ... \xrightarrow{r_l} t_{l+1}$ in $\mathcal{G}$, where $l\geq0$, a path hypothesis is defined as:
    \[H_{path}: P^{o}_{c}\left(\text{agg}(f_{\mathcal{P}}\mid M_{t_i}, \forall t_i \text{ on } \mathcal{P}) \right)\]
    where $o \in \{=, <>, >, <\}$; $c \in \mathbb{R}$ is a constant value; $P^{o}_{c}$ is the predicate in the format: equal, unequal, greater, or less than a value; $f_{\mathcal{P}}$ is any function of node and/or edge attributes on $\mathcal{P}$; 
    $\text{agg}(f_{\mathcal{P}}\mid M_{t_i}, \forall t_i \text{ on } \mathcal{P})$ is an aggregation function applying on the $f_{\mathcal{P}}$ of all paths whose nodes satisfy the corresponding modifiers.
\end{definition} 

    In the DBLP network defined previously, co-authorship can be represented as a path:
    \[\mathcal{P} = \texttt{author} \xrightarrow{\mathtt{Authorship}^{-1}} \texttt{paper} \xrightarrow{\mathtt{Authorship}} \texttt{author}\]
    ``The average citation of papers co-authored by Microsoft researchers is greater than 100'' can be expressed as:
    \[P^{>}_{100}\left(\textit{avg}(f_{\mathcal{P}}\mid \mathtt{author}[\mathtt{MSR}], \mathtt{paper}[],\mathtt{author}[\mathtt{MSR}]\right)\]
    where $f_{\mathcal{P}} = \mathtt{paper}[\mathtt{citation}]$, ${\tt MSR}$ stands for microsoft research.

When $l=0$, a path hypothesis is reduced to a \textbf{node hypothesis}. An example in DBLP, ``the average number of citations for conference papers is larger than 50'', can be expressed as: 
    \[P^{>}_{50}\left(\textit{avg}(\mathtt{paper}[\mathtt{citation}]\mid \mathtt{paper}[\mathtt{conference}])\right)\]

When $l=1$, we refer it as an \textbf{edge hypothesis}. For instance, the edge hypothesis, ``the average $\mathtt{FOS\_weight}$ of conference papers on data mining is larger than 0.5'', can be expressed as: 
\[P^{>}_{0.5}\left(\textit{avg}(\mathtt{WithDomain}[\mathtt{FOS\_Weight}]\mid \mathtt{paper}[\mathtt{C}], \mathtt{FOS}[\mathtt{DM}]\right)\]
{\tt C} (resp. {\tt DM}) stands for conference papers (resp. {\tt DM} data mining). %

\subsection{Problem Statement and Challenges}\label{def:problemStatement}
Given an attributed graph $\mathcal{G}$, our aim is to verify a node, edge, or path hypothesis $H$ and return true or false.
We formulate two objectives to address our problem: {\bf O1} - ensure hypothesis testing is as accurate as possible; {\bf O2} - minimize execution time.

When testing a hypothesis like ``the average number of citations for conference papers is greater than 50'', a conventional approach is to collect representative conference papers from the database, compute the average number of citations, and perform the statistical test. However, the entire graph may not be accessible, or collecting all relevant nodes, edges, or paths from $\mathcal{G}$ is time-consuming and impractical. Hence, we adopt graph sampling techniques to sample a subgraph $\mathcal{S}$ from $\mathcal{G}$, assuming a sampling budget $B$ that reflects the maximum size of $\mathcal{S}$. For simplicity, unless stated otherwise, sampling an edge or a node incurs the same unitary cost of one.

\textbf{A hypothesis estimator} computes an aggregate value for nodes, edges, or paths based on $H$. It directly impacts the accuracy of hypothesis testing. The challenge for hypothesis-agnostic samplers to achieve {\bf O1} is that they can miss relevant nodes, edges, or paths when $B$ is small. By examining the applicability of existing hypothesis-agnostic samplers, we observe that a large $B$ (almost the size of $\mathcal{G}$) is required to achieve an accuracy of one, which results in longer execution time and the challenge behind {\bf O2}. We conjecture that this is due to a lack of awareness of the underlying hypothesis during sampling. Hence, to address two challenges, we aim to design a {\em hypothesis-aware} sampler that preserves relevant nodes, edges, or paths in $S$. We also intend to show, both theoretically and empirically, that using our sampler, the hypothesis estimator on \(\mathcal{S}\) will {\em converge earlier} to its corresponding value in $\mathcal{G}$ as \(B\) increases.

\section{Sampling-based Hypothesis Testing}\label{sec:frm}
We propose a sampling-based hypothesis testing framework that supports both hypothesis-agnostic and hypothesis-aware graph samplers. After reviewing existing hypothesis-agnostic methods, we introduce our hypothesis-aware sampler, PHASE, in Section~\ref{frm:phase}. It achieves {\bf O1} by incorporating hypothesis awareness into the sampling, and {\bf O2} by ensuring earlier convergence of hypothesis estimators. Additionally, we implement two optimizations in $\text{PHASE}_{\text{opt}}$ to reduce the execution time of PHASE in Section ~\ref{frm:opt-phase}.

\subsection{Hypothesis-Agnostic Samplers}
Figure~\ref{fig:framework} illustrates our sampling-based hypothesis testing framework for testing node, edge, and path hypotheses on attributed graphs. It consists of two steps: (1) Sampling and (2) Hypothesis Testing. There are two workflows: hypothesis-agnostic, shown with dashed arrows, where $H$ is considered only in the hypothesis testing step, and hypothesis-aware, indicated by dash-dot-dash arrows, which requires $H$ during the sampling step.

In the hypothesis-agnostic workflow (dashed arrows), given $\mathcal{G}$ and $B$, a hypothesis-agnostic sampler is used in the sampling step to obtain $\mathcal{S}$. Existing hypothesis-agnostic samplers fall into three categories: node samplers that choose $B$ nodes from $\mathcal{G}$~\cite{stumpf2005subnets, DBLP:journals/corr/cs-NI-0103016}, edge samplers that choose $B$ edges from $\mathcal{G}$~\cite{DBLP:conf/networking/KrishnamurthyFCLCP05}, and random walk based samplers that pick edges and nodes by random walks~\cite{DBLP:conf/infocom/GjokaKBM10, DBLP:conf/sigmetrics/LeeXE12, DBLP:conf/kdd/LeskovecF06, DBLP:conf/icde/LiWL0LXL19}. 
In the hypothesis testing step, relevant nodes, edges, or paths are extracted from $\mathcal{S}$ for hypothesis testing. Finally, the acceptance result, p-value, and confidence interval are returned.

\subsection{Hypothesis-Aware Samplers}\label{frm:hypothesis-aware} 
\subsubsection{PHASE Algorithm}\label{frm:phase}
We introduce PHASE, our Path-\linebreak Hypothesis-Aware SamplEr (Algorithm~\ref{algo:ours}), for the sampling step. Since  $\mathcal{G}$ may contain both relevant and irrelevant nodes, edges, or paths for a given $H$, PHASE aims to preferentially sample those specified in $H$. This strategy resembles stratified sampling \cite{cochran1977sampling,neyman1992two}, where the target population is sampled at higher rates without bias.
PHASE can be integrated with any random walk based sampler. To clarify, we describe it using Frontier Sampler (\textit{FrontierS})~\cite{DBLP:conf/imc/RibeiroT10}, which picks a node from $m$ random seeds based on degree-proportional probability and performs a random walk by uniformly selecting a neighboring node. This process repeats until $B$ is reached.

\begin{algorithm}[t!]
    \small
    \caption{PHASE}
    \label{algo:ours}
    \begin{algorithmic}[1]
        \Statex \textbf{Input:} $\mathcal{G} = (\mathcal{V, E})$, $B$, $H$, $Q$
        \Statex \textbf{Output:} a sampled graph $\mathcal{S}$
        
        \State Initialize $L=(v_1, v_2, \ldots, v_m)$ with $m$ randomly chosen nodes
        \State $L_w=\text{AssignWeight}(L,H)$ \Comment{Assign $w_h$ to $x_1$ nodes and $w_l$ to others.}
        \State $\mathcal{V}_\mathcal{S}=\{\}$, $\mathcal{E}_\mathcal{S}=\{\}$
        \While{$B > m$}
            \State Normalize $L_w$
            \State Select $v\in L$ with probability $L_w$
            \State $N \gets N[v]$ \Comment{$N[v]$ is the set of neighbors of $v$.}
            \State $N_w=\text{AssignWeight}(N,H,{Q})$ \Comment{Assign weights according to {$Q$}.}
            \State Select $u \in N$ with the {normalized} $N_w$ 
            \State $\mathcal{V}_\mathcal{S}$.update($v,u$)
            \State Replace $v$ by $u$ in $L$ and update $L_w$ 
            \State $B=B-1$
        \EndWhile
        \State {\bf return} $\mathcal{S} = \{\mathcal{V}_\mathcal{S},\mathcal{E}_\mathcal{S}\}$
    \end{algorithmic}
\end{algorithm}

PHASE takes $\mathcal{G}$, $B$, $H$, $Q$ as inputs. For a node, edge, or path hypothesis, $H$ contains $\mathcal{P}$ with lengths zero, one, or more than one, respectively. $m$ seed nodes in $\mathcal{G}$ are randomly initialized and stored in a list $L$ (Line 1). This line increases the chance of picking relevant nodes, edges, and paths and ensures they are not locally clustered, preventing locality bias.
Each seed is assigned a weight in $L_w$ (Line 2) to guide the selection of seed nodes for random walks later. The weights are determined based on heuristics: nodes satisfying the first node modifier on $\mathcal{P}$ (denoted by $x_1$) receive a higher weight $w_{h}$, while others receive a lower weight $w_{l}$, where $w_{h} \geq w_{l} > 0$.

Lines 4-14 describe an iterative random walk to select nodes until $B$ is reached. During each iteration, the algorithm chooses a node $v$ from $L$ based on the probability distribution $L_w$ (Line 6). Next, it picks a neighbor $u$ of $v$ using a weighted selection process (Lines 7-9) ({\bf O1}). Specifically, in Line 8, we steer the random walks towards relevant nodes, edges, or paths. The transition probability matrices for node, edge, and path hypotheses are shown in Figures ~\ref{fig:transition_prob}a, ~\ref{fig:transition_prob}b, and ~\ref{fig:transition_prob}c, respectively.
For a node hypothesis, a higher weight \( w_h \) is assigned to \( x_1 \) regardless of the current node. For an edge hypothesis, \( w_h \) is given to \( x_2 \) when the current node is \( x_1 \), and given to \( x_1 \) in other cases. This increases the chance of sampling more relevant edges \( (x_1, x_2) \). The random walks are 1st-order in Figures ~\ref{fig:transition_prob}a and ~\ref{fig:transition_prob}b. Path hypotheses ($l=2$) have 2nd-order random walks (Figure ~\ref{fig:transition_prob}c), meaning the transition depends on the current and previous nodes. For example, given two nodes \( x_1\) and \(x_2\), \(x_3\) gets a higher weight \( w_h \). The consistent $w_h$ prevents sampling bias within the population. After adding nodes $u$ and $v$ to $\mathcal{V}_S$ (Line 10), the algorithm updates $L$ by replacing $v$ with $u$ (Lines 11) and decreases the sampling budget by one (Lines 12). This iteration continues until $B > m$. The resulting sampled graph $\mathcal{S}$ is the induced subgraph from $\mathcal{V}_S$.

\begin{figure}
\centering
\begin{tikzpicture}[baseline={(current bounding box.south)},
column 1/.style={anchor=base},
column 2/.style={anchor=base},
column 3/.style={anchor=base}]
   
  \matrix (left_matrix)
   [nodes={minimum size=6mm,anchor=center,font=\small}]
  {
    \node (left_matrix_1){}; & \node{$x_1$}; & \node {$y$}; \\ \hline
    \node {$x_1$}; & \node[color=red]{$w_h$}; & \node {$w_l$}; \\
    \node (left_matrix_7){$y$}; & \node[color=red]{$w_h$}; & \node {$w_l$}; \\
  };
    \draw (left_matrix_1.north east) -- (left_matrix_7.south east);
    \node[below=0.7cm of left_matrix,font=\small] {(a)};
    
    \matrix (mid_matrix) [right=0.1cm of left_matrix,nodes={minimum size=6mm,anchor=center},font=\small]
  {
   \node (mid_matrix_1){}; & \node{$x_1$}; & \node{$x_2$} ; & \node {$y$}; \\ \hline
    \node {$x_1$}; & \node{$w_{l}$}; & \node[color=red]{$w_{h}$}; & \node {$w_{l}$}; \\
     \node {$x_2$}; & \node[color=red]{$w_h$}; & \node{$w_l$}; & \node {$w_l$}; \\
    \node (mid_matrix_7){$y$}; & \node[color=red]{$w_h$}; & \node{$w_l$}; & \node {$w_l$}; \\
  };
  \draw (mid_matrix_1.north east) -- (mid_matrix_7.south east);
  \node[below=0.4cm of mid_matrix,font=\small] {(b)};
  
  \matrix (right_matrix) [right=0.1cm of mid_matrix,nodes={minimum size=6mm,anchor=center},font=\small]
  {
   \node (right_matrix_1)[minimum width=8mm]{}; & \node{$x_1$}; & \node{$x_2$}; & \node{$x_3$}; & \node {$y$}; \\ \hline
    \node [minimum width=8mm]{$x_1, x_1$}; & \node{$w_l$}; & \node[color=red]{$w_h$}; & \node{$w_l$}; & \node {$w_l$}; \\
    \node [minimum width=8mm]{$x_1, x_2$}; & \node{$w_l$}; & \node{$w_l$}; & \node[color=red]{$w_h$}; & \node {$w_l$}; \\
    \node [minimum width=8mm]{$x_1, x_3$}; & \node[color=red]{$w_h$}; & \node{$w_l$}; & \node{$w_l$}; & \node {$w_l$}; \\
    \node (right_matrix_7)[minimum width=8mm]{$x_1, y$}; & \node[color=red]{$w_h$}; & \node{$w_l$}; & \node{$w_l$}; & \node {$w_l$}; \\
  };
  \draw (right_matrix_1.north east) -- (right_matrix_7.south east);
  \node[below=0.1cm of right_matrix,font=\small] {(c)};
\end{tikzpicture}


\caption{Transition probability matrices \({Q} \) for (a) node, (b) edge, and (c) path $(l=2)$ hypotheses (up to the first four rows). $x_i$ represents nodes in $\mathcal{G}$ satisfying the $i$-th node modifier on $\mathcal{P}$ and $y$ represents other nodes.}

\label{fig:transition_prob}
\end{figure}

\subsubsection{$\text{PHASE}_{\text{opt}}$ Algorithm}\label{frm:opt-phase}
We further propose optimizations to reduce execution time based on heuristics. Algorithm~\ref{algo:phase_opt} introduces two lines that replace line 7 in Algorithm~\ref{algo:ours}. In line 1, we employ a non-backtracking approach to avoid selecting previously visited nodes during sampling~\cite{DBLP:conf/sigmetrics/LeeXE12}, preventing cycles and unnecessary revisits. Additionally, in densely connected graphs, computing neighbor weights in line 8 of Algorithm~\ref{algo:ours} can be time-consuming due to the potentially large number of neighbors. To address this, line 2 in Algorithm~\ref{algo:phase_opt} introduces random sampling of a subset of $\min\{|N'|,n\}$ neighbors as candidates, where \( n \) is a parameter. This can effectively reduce computation time. But it may result in some accuracy loss when \( n \) is significantly smaller than \(|N'|\).

\begin{algorithm}
    \caption{$\text{PHASE}_{\text{opt}}$ \\ (showing only 2 lines that replace line 7 in Algorithm ~\ref{algo:ours}}
    \small
    \begin{algorithmic}[1]
        \State $N' \gets N[v] - \mathcal{V_S}$  (Optim 1)
        \State $N \gets $ Select $\text{min}\{ |N'|, n \} \text{ nodes randomly from } N'$ (Optim 2)
    \end{algorithmic}
    \label{algo:phase_opt}
\end{algorithm} 

\stitle{Complexity.} On average, each node has $2|\mathcal{E}|/|\mathcal{V}|$ neighbors to be examined by PHASE, resulting in a time complexity of $O(B \times 2|\mathcal{E}|/|\mathcal{V}|)$. On the other hand, by constraining the number of neighbors, $\text{PHASE}_{\text{opt}}$ achieves a time complexity of $O(B)$.

\subsubsection{Convergence of Hypothesis Estimators}\label{frm:convergence}

We demonstrate the convergence of hypothesis estimators, which ensures the convergence of hypothesis testing accuracy to one, for all sampling methods. Then, we show our proposed sampler achieves earlier and smoother convergence ({\bf O2}).

We will focus on scenarios where $\text{agg}$ is an average function to construct and prove the convergence of hypothesis estimators. Estimators for other aggregation functions, such as maximum and minimum, can be derived analogously.

For a path hypothesis ($l\geq0$), $\text{avg}(f_{\mathcal{P}}\mid M_{t_i}, \forall t_i \text{ on } \mathcal{P})$ has a primary subject $f_{\mathcal{P}}$. Let $\mathcal{P}^*$ be all relevant paths in $\mathcal{G}$, and $\mathcal{P_G}$ be all paths with the same length as $\mathcal{P}$ in $\mathcal{G}$. The mean value of the path hypothesis, $\theta_{path}$, is 
\begin{equation}
\theta_{path} = \frac{1}{|\mathcal{P}^*|}\sum_{\forall \mathcal{P} \in \mathcal{P_G}} T(\mathcal{P})
\label{eq:7}
\end{equation}
where $T(\mathcal{P}) = f_{\mathcal{P}} \times \mathbbm{1}_{M_{t_i}\subseteq \mathcal{L}_{\phi(t_i)} \forall t_i \text{ on } \mathcal{P}}$. Replacing $\mathcal{G}$ with the sampled graph $\mathcal{S}$, the estimator for $\theta_{path}$ on $\mathcal{S}$ is
\begin{equation}
\hat{\theta}_{path} = \frac{1}{|\mathcal{P}^*\cap\mathcal{P_{S}}|}\sum_{\forall \mathcal{P} \in \mathcal{P_{S}}} T(\mathcal{P})
\label{eq:8}
\end{equation}
When $l=0$ (resp. $l=1$), we name the corresponding mean value $\theta_{node}$ (resp. $\theta_{edge}$) and estimator $\hat{\theta}_{node}$ (resp. $\hat{\theta}_{edge}$).

Some sampling methods, including random edge sampler, simple random walk, and \textit{FrontierS}, obey the Strong Law of Large Numbers (SLLN) ~\cite{DBLP:conf/imc/RibeiroT10}. Their $\hat{\theta}_{node}$ and $\hat{\theta}_{edge}$ are asymptotically unbiased estimators of $\theta_{node}$ and $\theta_{edge}$, respectively, when $B \to \infty$.

\begin{theorem}[SLLN]
For any function $f$, where \\ $\sum_{(u,v)\in \mathcal{E}}|f(u,v)|<\infty$,
\[\lim_{B\to\infty} \frac{1}{B}\sum^{B}_{i=1}f(u_i,v_i) \to \frac{1}{|E|}\sum_{\forall (u,v)\in\mathcal{E}}f(u,v)\]
almost surely, i.e. the event occurs with probability one.
\label{thm:SLLN}
\end{theorem}

In general scenarios with any path hypotheses ($l \geq 0$) and sampler, SLLN may not apply, and the estimator may not be asymptotically unbiased. However, \(\hat{\theta}_{\text{path}}\) still converges to \(\theta_{\text{path}}\) due to Finite Population Correction (FPC), a statistical adjustment made when sampling without replacement from a finite population~\cite{bondy1976standard}. As \(B\) increases, \( \text{FPC} = \frac{|\mathcal{P}^*| - \mathcal{P_S}}{|\mathcal{P}^*| - 1} \) approaches zero, which ensures the convergence of the hypothesis estimator to its ground truth.



\begin{figure}[t!]
    \centering
    \begin{subfigure}[b]{0.49\columnwidth}
        \centering
        \includegraphics[width=\columnwidth]{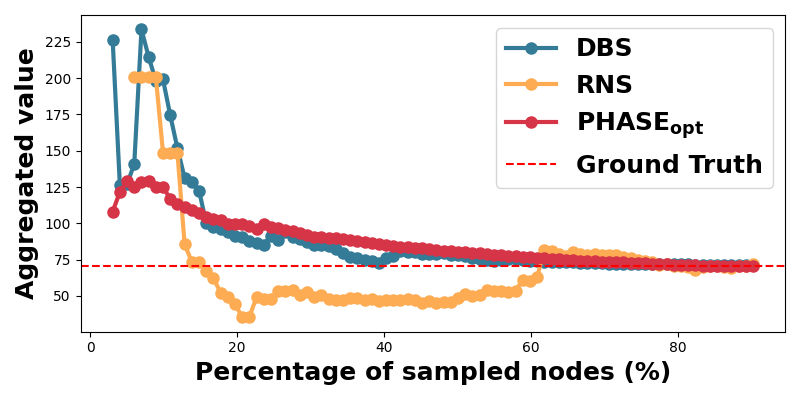}
        \label{fig:dblp-convergence}
    \end{subfigure}
    \hfill
    \begin{subfigure}[b]{0.49\columnwidth}
        \centering
        \includegraphics[width=\columnwidth]{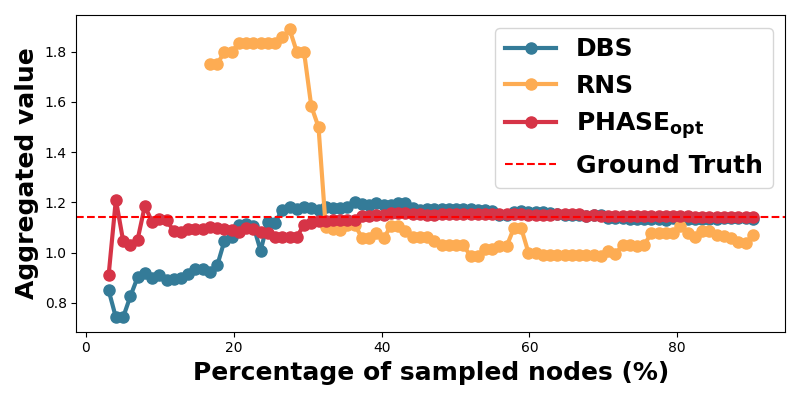}
        \label{fig:yelp-convergence}
    \end{subfigure}
    \caption{The convergence of hypothesis estimator for two path hypotheses: DB-P3 (left) and YP-P3 (right).}
    \label{fig:convergence}
\end{figure}

While all samplers ensure the convergence of the hypothesis estimator, our proposed samplers achieve earlier convergence than other methods. This is due to their higher probability of selecting relevant nodes, edges, and paths. For instance, when $l=2$, the probability to pick nodes $(x_1, x_2, x_3)$ is:
\begin{align*}
P(x_1, x_2, x_3) &= P(x_1) \times (P(x_2 \mid x_1) \times A(x_2 \mid x_1)) \\
                 &\quad \times (P(x_3 \mid x_1, x_2) \times A(x_3 \mid x_1, x_2))
\end{align*}
where \(A(x_i \mid x_{i-1})\) indicates the accessibility (0 or 1) of node \(x_i\) from \(x_{i-1}\). Assuming node \(x_i\) has \(n_i\) neighbors with \(k_i\) of type \(x_{i+1}\) and there are $d$ $x_1$ in $m$ seed nodes, where $d\geq 0$ and $k_i \geq 0$:
\begin{align*}
P_{\text{PHASE}}(x_1, x_2, x_3) &= \frac{d\cdot w_h}{d\cdot w_h+(m-d)\cdot w_l} \\
                                &\quad \times \left(\frac{w_h}{k_1 \cdot w_h + (n_1 - k_1) \cdot w_l} \times A(x_2 \mid x_1)\right) \\
                                &\quad \times \left(\frac{w_h}{k_2 \cdot w_h + (n_2 - k_2) \cdot w_l} \times A(x_3 \mid x_1, x_2)\right)
\end{align*}
Random node and edge samplers have the lowest probability of picking $(x_1, x_2, x_3)$. Random walk based samplers have a transitional probability from $x_i$ of $\frac{1}{n_i} \leq \frac{w_h}{k_i \cdot w_h + (n_i - k_i)\cdot w_l}$, particularly when \( k_i \) is 
large. With a higher probability to pick $(x_1, x_2, x_3)$, PHASE and $\text{PHASE}_{\text{opt}}$ have earlier convergence of the hypothesis estimator than existing hypothesis-agnostic samplers. Also, as shown in Figure \ref{fig:convergence}, the empirical results of two path hypotheses, whose details are in Table~\ref{tab:hypo}, align with the claim.

\section{Experiments}\label{sec:exps}
The goal of our experiments is twofold: 1) test the effectiveness of optimizations in $\text{PHASE}_{\text{opt}}$ compared to PHASE (Section ~\ref{expx:PHASEvsOPT}), and 2) compare hypothesis-agnostic and hypothesis-aware samplers in terms of test significance (Section ~\ref{expx:significance}), accuracy (Section ~\ref{expx:accuracy}), and time (Section ~\ref{expx:time}). 

\subsection{Experimental Setup}\label{subsec:dataset}
\begin{table}[t]
\caption{Statistics of datasets}
\centering
\scriptsize
\resizebox{\columnwidth}{!}{%
\begin{tabular}{|c|c|c|c|>{\centering\arraybackslash}m{0.15\columnwidth}|>{\centering\arraybackslash}m{0.15\columnwidth}|}

\hline
\textbf{Dataset} & \textbf{\#(Nodes)}  & \textbf{\#(Edges)} & \textbf{Density} &\textbf{\#(Node Types)} & \textbf{\#(Edge Types)} \\ \hline
MovieLens& 9,705      & 996,656   & 1.06e-02 & 2 & 1 \\ \hline
DBLP     & 1,623,013  & 11,040,170 & 4.19e-06 & 4 & 4 \\ \hline
Yelp     & 2,136,118  & 6,743,879 & 1.48e-06 & 2 & 1 \\ \hline
\end{tabular}
}
\label{tab:dataset_summary}
\end{table}

\stitle{Datasets.} We use three datasets~\cite{DBLP:journals/tiis/HarperK16, DBLP:conf/kdd/TangZYLZS08, yelp_dataset}, extracted from real attributed networks: MovieLens, DBLP, and Yelp. Table~\ref{tab:dataset_summary} shows the statistics of the datasets.

\stitle{Samplers}
We compare $\text{PHASE}_{\text{opt}}$ with 11 existing samplers: 
\begin{itemize}[leftmargin=*]
    \item Node samplers: Random Node Sampler (\textit{RNS})~\cite{stumpf2005subnets}, Degree-Based Sampler (\textit{DBS}) ~\cite{DBLP:journals/corr/cs-NI-0103016}
    \item Edge samplers: Random Edge Sampler (\textit{RES}) ~\cite{DBLP:conf/networking/KrishnamurthyFCLCP05}
    \item Random walk based samplers: Simple Random Walk (\textit{SRW}) ~\cite{DBLP:conf/infocom/GjokaKBM10}, Frontier Sampler (\textit{FrontierS}) ~\cite{DBLP:conf/imc/RibeiroT10}, Non-Backtracking Random Walk (\textit{NBRW}) ~\cite{DBLP:conf/sigmetrics/LeeXE12}, Random Walk with Restarter (\textit{RWR}) ~\cite{DBLP:conf/kdd/LeskovecF06}, Metropolis-Hastings Random Walk (\textit{MHRW}) ~\cite{DBLP:journals/ton/StutzbachRDSW09}, Snaw Ball Sampler (\textit{SBS}) ~\cite{goodman1961snowball}, Forest Fire Sampler (\textit{FFS}) ~\cite{DBLP:conf/kdd/LeskovecF06}, Shortest Path Sampler (\textit{ShortestPathS}) ~\cite{rezvanian2015sampling}
\end{itemize}

\stitle{Hypotheses.}
For each dataset (e.g., DB) and hypothesis type (e.g., N for node hypotheses), we chose three example hypotheses (e.g., DB-N1) in the experiment. Due to the page limit, examples for DBLP and Yelp are shown in Table \ref{tab:hypo}. These hypotheses are chosen based on context and difficulty. Based on the context of the datasets, the subject of interest and the constant in the hypothesis should reflect meaningful requests from real users. The difficulty is related to the path length and the number of relevant nodes, edges, or paths, as indicated in the third column of Table \ref{tab:hypo}. The longer the path or the fewer the relevant nodes, edges, or paths, the more difficult it becomes to sample them from $\mathcal{G}$ for accurate hypothesis testing. We use the following path hypotheses of DBLP with lengths of three and four in the experiment: ``the average citation of two papers, each authored by a Microsoft researcher and citing each other, $>$ 50'' and ``the average citation of two conference papers, each authored by a Microsoft researcher, $>$ 50''.

\begin{table}[h!]
\caption{Examples of hypotheses for DBLP and Yelp}
\label{tab:hypo}
\resizebox{\columnwidth}{!}{%
\fontsize{7}{9}\selectfont
\begin{tabular}{|m{0.16\columnwidth}|m{0.6\columnwidth}|m{0.23\columnwidth}|}
\hline
\textbf{Hypothesis \newline Type}     & \multicolumn{1}{c|}{\textbf{Example}}                                                                            & \textbf{Relevant nodes, \newline edges, paths} \\ \hline

\multicolumn{3}{|c|}{\textbf{DBLP}} \\ \hline

\multirow{5}{*}{Node} & \textbf{DB-N1:} The avg citation of papers published as journals $>$  20                                                            & 199205 (easy) 
                      \\ \cline{2-3} 
                      & \textbf{DB-N2:} The avg   citation of papers in conferences in 2010 $>$ 10                            & 31566 (medium)
                      \\ \cline{2-3} 
                      & \textbf{DB-N3:} The avg   citation of papers published in Journal in 2017 $>$ 10                                & 1588 (hard)
                      \\ \hline
\multirow{5}{*}{Edge} & \textbf{DB-E1:} The avg weight of conference   papers on data mining $>$ 0.5                                    & 44925 (easy)
                      \\ \cline{2-3} 
                      & \textbf{DB-E2:} The avg   weight of journal papers on data mining $>$ 0.5                                       & 13400 (medium)
                      \\ \cline{2-3} 
                      & \textbf{DB-E3:} The avg   weight of papers on telecommunications network $>$ 0.5                                & 2510 (hard)
                      \\ \hline
\multirow{5}{*}{Path} & \textbf{DB-P1:} The avg weight of papers by China's institutes on data mining $>$ 0.5              & 17671 (easy)                       \\ \cline{2-3} 
                      & \textbf{DB-P2:} The avg   citation of papers co-authored by authors in Peking and China's institutions   $>$ 50 & 7065 (medium)                        \\ \cline{2-3} 
                      & \textbf{DB-P3:} The avg   citation of conference papers by Microsoft Researchers $>$ 50                 & 3217 (hard)                                    \\ \hline
                    
\multicolumn{3}{|c|}{\textbf{Yelp}} \\ \hline
\multirow{5}{*}{Node} & \textbf{YP-N1:} The avg reviews given by users   with high popularity $>$ 200                                   & 112043 (easy)
\\ \cline{2-3} 
                      & \textbf{YP-N2:} The avg   stars given by users who have low prolificacy and medium popularity $>$ 3             & 16429 (medium)
                      \\ \cline{2-3} 
                      & \textbf{YP-N3:} The avg   number of reviews of business in Illinois $>$ 4                                      & 2144 (hard)
                      \\ \hline
\multirow{4}{*}{Edge} & \textbf{YP-E1:} The avg ratings of fast food $>$ 4                                                            & 224536 (easy)
\\ \cline{2-3} 
                      & \textbf{YP-E2:} The avg   ratings of furniture stores $>$ 3                                                     & 33040 (medium)
                      \\ \cline{2-3} 
                      & \textbf{YP-E3:} The avg   percentage of useful reviews given by useful writers to Illinois businesses   $>$ 0.5 & 4242 (hard)
                      \\ \hline
\multirow{6}{*}{Path} & \textbf{YP-P1:} The avg rating difference on path [business in FL - high popularity user - business in LA] $>$ 0.5   &   615174 (easy)                  \\ \cline{2-3} 
                      &  \textbf{YP-P2:} The avg rating difference on path [business in LA - high popularity user - business in IL] $>$ 0.5   &  15542 (medium)                \\ \cline{2-3} 
                      &  \textbf{YP-P3:} The avg rating difference on path [business in LA - medium popularity user - business in AB] $>$ 0.5   &  1080 (hard)         \\ \hline
\end{tabular}%
}
\end{table}

\stitle{Parameter Choice.} 
We determine optimal settings for $m$, $n$, $w_h$, and $w_l$ through a grid search, with $m$ and $n$ ranging from 10 to 200, and $w_h$ and $w_l$ from 0.1 to 20. We set $m=50$ to maintain the path-preserving ability of multi-dimensional random walks, and $n=30$ to strike a balance between accuracy and time efficiency. $w_h=10$ and $w_l=0.1$ help regulate the prioritization towards sampling relevant nodes, edges, or paths. The sampling budget $B$ is maintained as a proportion of the total number of nodes in $\mathcal{G}$ for all samplers. In real deployment, we recommend iteratively increasing \( B \) until the accuracy stabilizes at a high threshold (e.g., 0.9) based on the average from 30 samples to determine the optimal $B$.
As for existing hypothesis-agnostic samplers, we use the best parameters in their respective settings. We report an average of 30 runs for every evaluation measure. 

\subsection{Evaluation Measures}\label{expx:evaMethod}
\stitle{Accuracy.} 
Accuracy at a specific sampling proportion reflects the effectiveness of a sampling method. It measures the number of matched hypothesis testing results on $\mathcal{G}$ and $\mathcal{S}$:
\[\text{Accuracy} = \frac{1}{k}\sum_k \mathbbm{1}_{H(\mathcal{G}) == H(\mathcal{S})}\]
\noindent
where $H(\mathcal{G})$ and $H(\mathcal{S})$ return 0 (false) or 1 (true). 

\stitle{Time.} We measure the total execution time, including the sampling time, the time taken to extract relevant information from $\mathcal{S}$ for hypothesis testing, and hypothesis testing time.

\begin{table*}[tb]
\caption{The accuracy of 11 existing samplers and $\text{PHASE}_{\text{opt}}$ on three datasets and three types of hypothesis.}
\scriptsize
\resizebox{\textwidth}{!}{%
\begin{tabular}{|c|>{\centering\arraybackslash}m{0.05\textwidth}|>{\centering\arraybackslash}m{0.08\textwidth}|c|c|c|c|c|c|c|c|c|c|c|c|c|}
\hline
Dataset                     & Hypothesis Type & Sampling   Proportion (\%) & $\text{PHASE}_{\text{opt}}$ & {PHASE}        & RES  & RNS  & DBS  & SRW                         & NBRW       & RWR & MHRW & ShortestPathS & FrontierS & FFS  & SBS  \\ \hline
                            & Node            & 1                          &  \underline{0.9} & {\textbf{1}}   & 0.89 & 0.83 & 0.86 & 0.88                        & 0.89       & 0.87        & 0.89  & 0.86          & 0.84      & 0.83 & 0.56 \\ \cline{2-16} 
                            & Edge            & 2.5                        & 0.98    & {\textbf{1}}        & 0.68 & 0.77 & 0.98 & \underline{0.99}        & \textbf{1} & \underline{0.99}        & 0.98  & \underline{0.99}          & 0.73      & 0.93 & 0.91 \\ \cline{2-16} 
\multirow{-3}{*}{MovieLens} & Path            & 5                          & \underline{0.99} & {\textbf{1}}   & 0.1  & 0.88 & 0.83 & 0.82                        & 0.89       & 0.91        & 0.98  & 0.95          & 0.38      & 0.85 & 0.62 \\ \hline
                            & Node            & 0.2                        & \underline{0.96}  & {\textbf{1}} & 0.48 & 0.87 & 0.93 & 0.92                        & 0.91       & 0.9         & 0.94  & 0.92          & 0.92      & 0.94 & 0.88 \\ \cline{2-16} 
                            & Edge            & 0.2                        & \underline{0.76}  & {\textbf{0.79}} & 0.48 & 0    & 0.71 & 0.73 & 0.69       & 0.7         & 0.29  & 0.69          & 0.63      & 0.7  & 0.42 \\ \cline{2-16} 
\multirow{-3}{*}{DBLP}      & Path            & 0.2                        & \textbf{0.89} & {\underline{0.88}} & 0    & 0    & 0    & 0.26                        & 0.3        & 0.32        & 0.18  & 0.33          & 0.043     & 0.3  & 0.12 \\ \hline
                            & Node            & 0.1                        & \underline{0.99} & {\textbf{1}} & 0.65 & 0.77 & 0.61 & 0.69                        & 0.69       & 0.69        & 0.77  & 0.7           & 0.64      & 0.66 & 0.48 \\ \cline{2-16} 
                            & Edge            & 1                          & \textbf{1} & {\textbf{1}}   & 0.73 & 0.54 & 0.76 & 0.79                        & \underline{0.91}       & 0.87        & 0.84  & 0.76          & 0.77      & 0.79 & 0.71 \\ \cline{2-16} 
\multirow{-3}{*}{Yelp}      & Path            & 1                          & \underline{0.99}& {\textbf{1}} & 0.11 & 0.05 & 0.79 & 0.78                        & 0.78       & 0.72        & 0.8   & \underline{0.99} & 0.42      & 0.67 & 0.42 \\ \hline
\end{tabular}%
}
\label{tab:acc-table}
\end{table*}

\begin{table*}[tb]
\caption{The execution time (sec) of 11 existing samplers and $\text{PHASE}_{\text{opt}}$ on three datasets and three types of hypothesis.}
\scriptsize
\resizebox{\textwidth}{!}{%
\begin{tabular}{|c|>{\centering\arraybackslash}m{0.05\textwidth}|>{\centering\arraybackslash}m{0.08\textwidth}|c|c|c|c|c|c|c|c|c|c|c|c|c|}
\hline
Dataset                    & Hypothesis Type & Sampling Proportion (\%) & $\text{PHASE}_{\text{opt}}$ & {PHASE}& RES   & RNS            & DBS   & SRW   & NBRW          & RWR   & MHRWS & ShortestPathS & FrontierS & FFS   & SBS   \\ \hline
\multirow{3}{*}{MovieLens} & Node            & 1                        & 0.083 &{0.41}     & 0.99  & \textbf{0.023} & 0.077 & 0.057 & 0.06          & 0.05  & 0.06  & 0.063         & 0.083     & 0.067 & 0.047 \\ \cline{2-16} 
                           & Edge            & 2.5                      & 0.45 & {2.07}     & 0.99  & \textbf{0.11}  & 0.36  & 0.31  & 0.36          & 0.33  & 0.29  & 0.26          & 0.33      & 0.35  & 0.31  \\ \cline{2-16} 
                           & Path            & 5                        & 4.92 &{14.80}      & 1.03  & \textbf{0.34}  & 4.32  & 4.53  & 4.55          & 4.76  & 3.10  & 0.95          & 0.38      & 3.87  & 3.08  \\ \hline
\multirow{3}{*}{DBLP}      & Node            & 0.2                      & 5.56 &{418.07}      & 18.70 & \textbf{0.55}  & 6.98  & 7.98  & 8.22          & 9.67  & 1.66  & 31.32         & 10.48     & 5.73  & 3.03  \\ \cline{2-16} 
                           & Edge            & 0.2                      & 8.76 & {414.47}     & 22.57 & \textbf{0.90}  & 10.07 & 14.76 & 12.89         & 12.32 & 3.53  & 33.91         & 14.46     & 8.87  & 5.27  \\ \cline{2-16} 
                           & Path            & 0.2                      & 5.44 & {236.82}      & 19.61 & \textbf{0.71}  & 6.87  & 8.24  & 8.39          & 8.13  & 2.17  & 31.13         & 9.37      & 5.33  & 3.01  \\ \hline
\multirow{3}{*}{Yelp}      & Node            & 0.1                      & 2.56 &{6.10}     & 13.84 & 1.02           & 6.81  & 1.19  & \textbf{0.96} & 1.16  & 1.12  & 1.84          & 1.68      & 1.02  & 6.58  \\ \cline{2-16} 
                           & Edge            & 1                        & 19.42 &{67.55}     & 16.88 & \textbf{2.01}  & 13.61 & 8.58  & 8.97          & 9.17  & 6.93  & 30.23         & 10.22     & 10.27 & 6.03  \\ \cline{2-16} 
                           & Path            & 1                        & 56.97 &{130.3}     & 14.35 & \textbf{1.53}  & 23.48 & 15.97 & 16.61         & 19.03 & 7.35  & 37.81         & 9.95      & 25.61 & 19.94 \\ \hline
\end{tabular}%
}
\label{tab:time-table}
\end{table*}

\subsection{PHASE vs $\text{PHASE}_{\text{opt}}$}\label{expx:PHASEvsOPT}
We evaluate $\text{PHASE}_{\text{opt}}$'s optimizations compared to PHASE in Table \ref{tab:acc-table} and \ref{tab:time-table}. For DBLP, $\text{PHASE}_{\text{opt}}$ is at least 43 times faster than PHASE with minimal accuracy loss (under 4\%). Consequently, we use $\text{PHASE}_{\text{opt}}$ exclusively in further experiments and discussions.


\subsection{Significance}\label{expx:significance}
\begin{figure}[!t]
    \centering
    \begin{subfigure}[b]{0.49\columnwidth}
        \centering
        \includegraphics[width=\columnwidth]{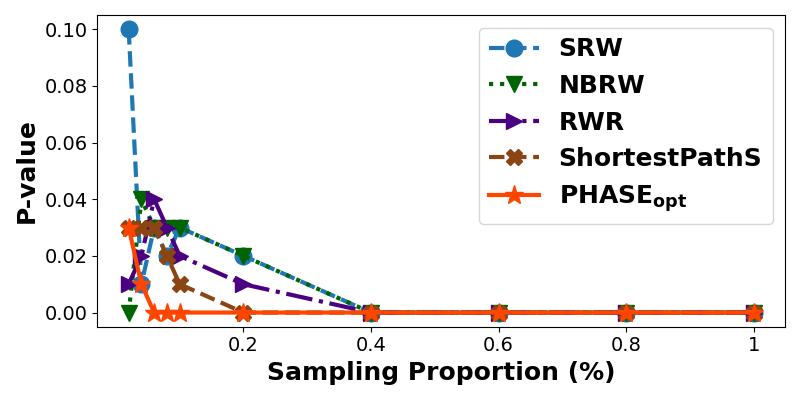}
        \caption{P-value Plot}
        \label{fig:dblp-p-value-path-1}
    \end{subfigure}
    \hfill
    \begin{subfigure}[b]{0.49\columnwidth}
        \centering
        \includegraphics[width=\columnwidth]{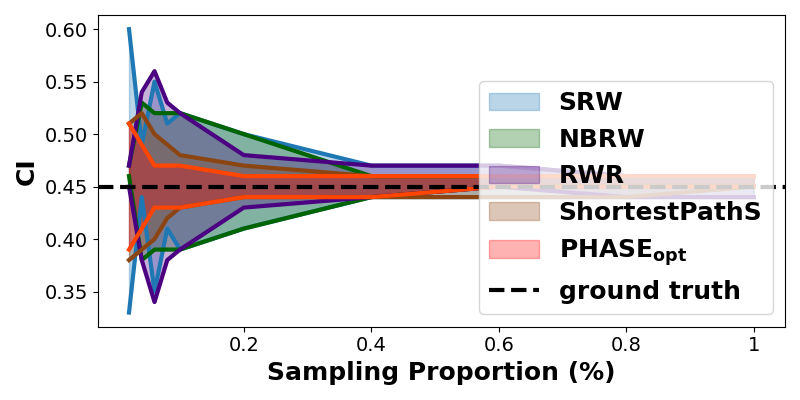}
        \caption{CI Plot}
        \label{fig:dblp-ci-path-1}
    \end{subfigure}
    \caption{DBLP p-value and CI plot for the hypothesis DB-P1}
    \label{fig:p-value_ci}
\end{figure}

To assess statistical significance and estimation precision, we evaluate p-value trends and confidence intervals (CIs) as $B$ increases using DB-P3, as shown in Figure ~\ref{fig:p-value_ci}. 
Similar trends are observed for other hypotheses.
$\text{PHASE}_{\text{opt}}$ consistently maintains p-values below the significance level (e.g., 0.05), showing 
stronger evidence against the null hypothesis as $B$ increases. It also exhibits the narrowest CI at all $B$, reflecting the highest precision among other samplers. 



\begin{figure}[!t]
    \centering
    \begin{subfigure}[b]{0.49\columnwidth}
        \centering
        \includegraphics[width=\columnwidth]{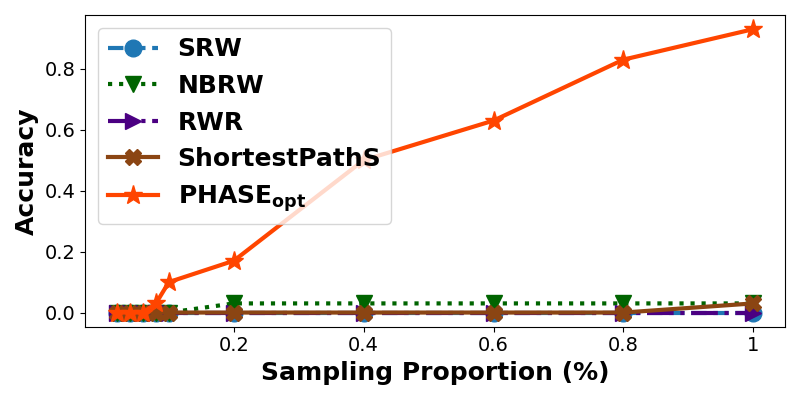}
        \caption{Accuracy $(l=3)$}
        \label{fig:dblp-APPA-acc}
    \end{subfigure}
    \hfill
    \begin{subfigure}[b]{0.49\columnwidth}
        \centering
        \includegraphics[width=\columnwidth]{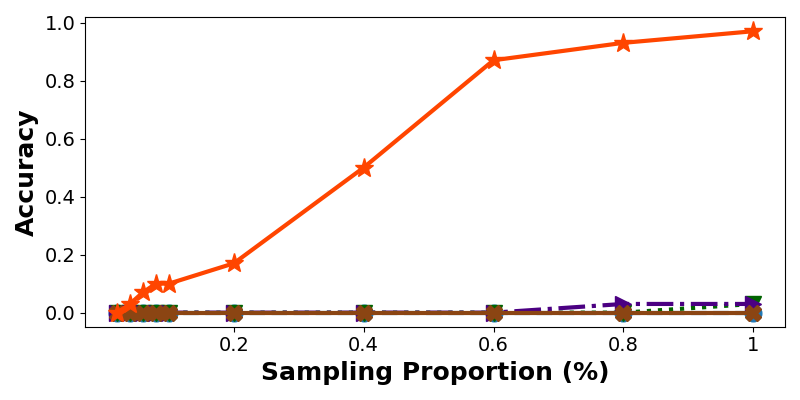}
        \caption{Accuracy $(l=4)$}
        \label{fig:dblp-APVPA-acc}
    \end{subfigure}
    \centering
    \begin{subfigure}[b]{0.49\columnwidth}
        \centering
        \includegraphics[width=\columnwidth]{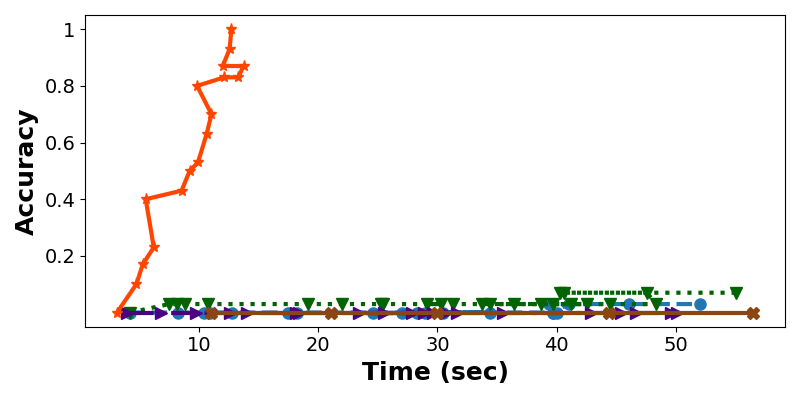}
        \caption{Time-Accuracy $(l=3)$}
        \label{fig:dblp-APPA-time}
    \end{subfigure}
    \hfill
    \begin{subfigure}[b]{0.49\columnwidth}
        \centering
        \includegraphics[width=\columnwidth]{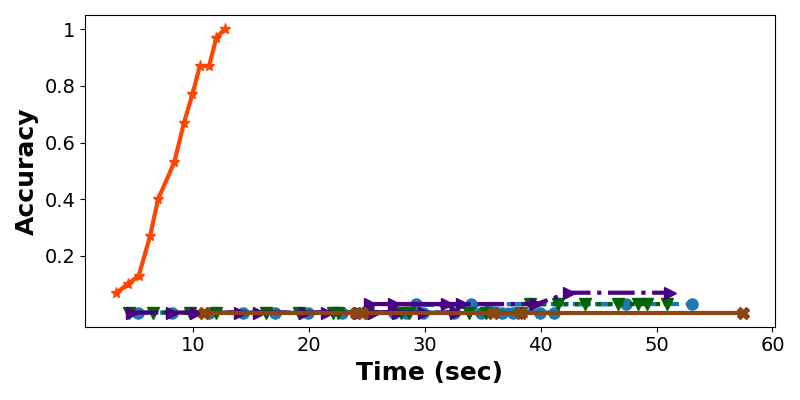}
        \caption{Time-Accuracy $(l=4)$}
        \label{fig:dblp-APVPA-time}
    \end{subfigure}
    \caption{Accuracy (a, b) and time-accuracy (c, d) performance for DBLP path hypotheses.}
    \label{fig:pathlength}
\end{figure}

\begin{figure*}[t!]
    \centering
    \begin{subfigure}[b]{0.24\linewidth}
        \centering
        \includegraphics[width=\linewidth]{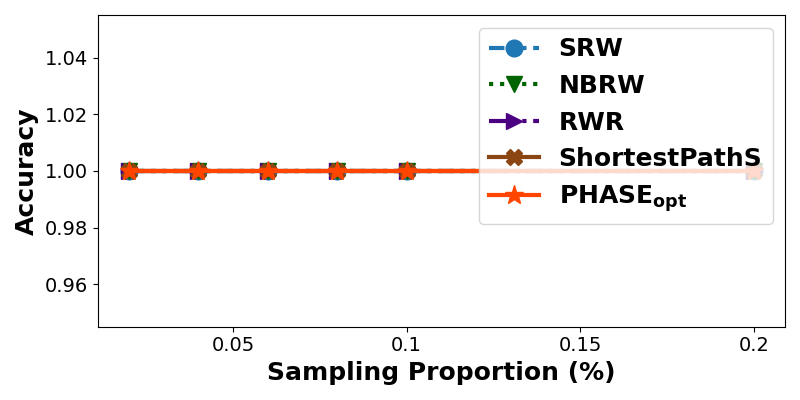}
        \caption{DB-N1 (easy)}
        \label{fig:dblp-acc-node-1}
    \end{subfigure}
    \hfill
    \begin{subfigure}[b]{0.24\linewidth}
        \centering
        \includegraphics[width=\linewidth]{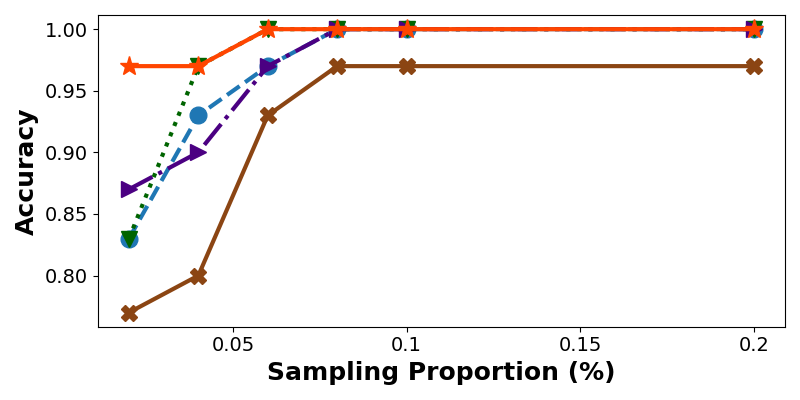}
        \caption{DB-N2 (medium)}
        \label{fig:dblp-acc-node-2}
    \end{subfigure}
    \hfill
    \begin{subfigure}[b]{0.24\linewidth}
        \centering
        \includegraphics[width=\linewidth]{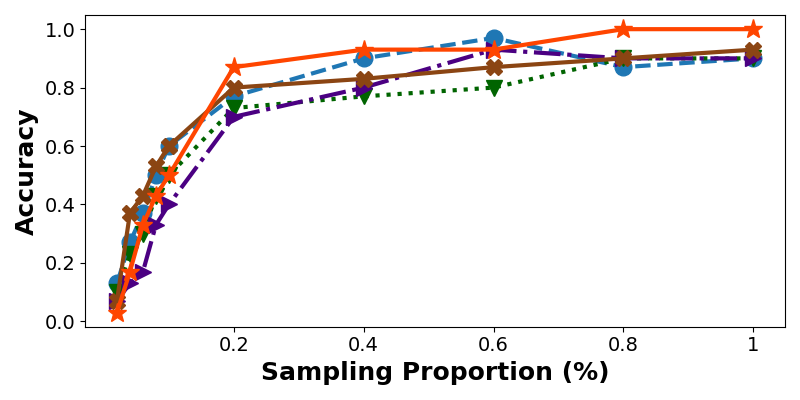}
        \caption{DB-N3 (hard)}
        \label{fig:dblp-acc-node-3}
    \end{subfigure}
    \hfill
    \begin{subfigure}[b]{0.24\linewidth}
        \centering
        \includegraphics[width=\linewidth]{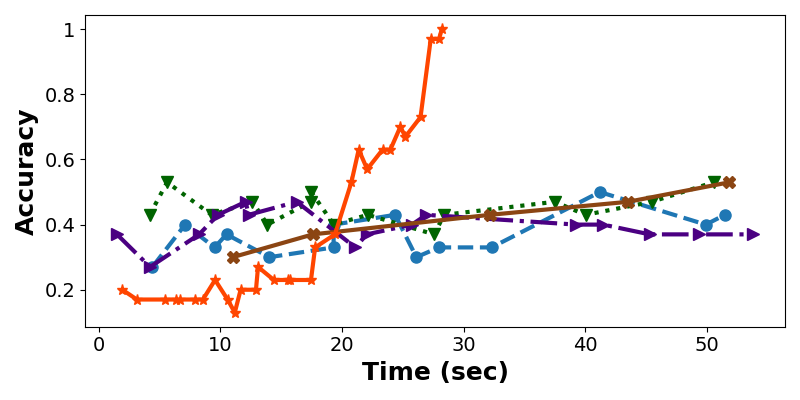}
        \caption{DB-N3 (hard)}
        \label{fig:dblp-time-node-1}
    \end{subfigure}

    \begin{subfigure}[b]{0.24\linewidth}
        \centering
        \includegraphics[width=\linewidth]{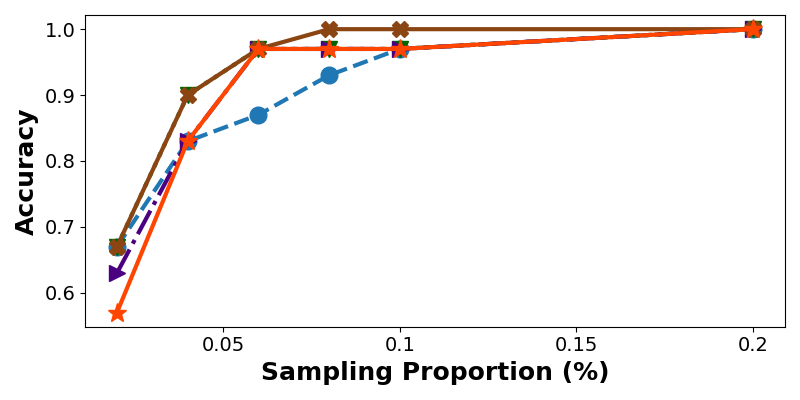}
        \caption{DB-E1 (easy)}
        \label{fig:dblp-acc-edge-1}
    \end{subfigure}
    \hfill
    \begin{subfigure}[b]{0.24\linewidth}
        \centering
        \includegraphics[width=\linewidth]{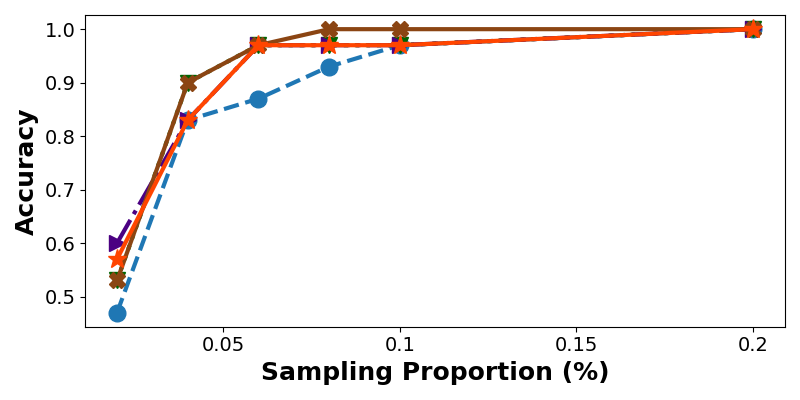}
        \caption{DB-E2 (medium)}
        \label{fig:dblp-acc-edge-2}
    \end{subfigure}
    \hfill
    \begin{subfigure}[b]{0.24\linewidth}
        \centering
        \includegraphics[width=\linewidth]{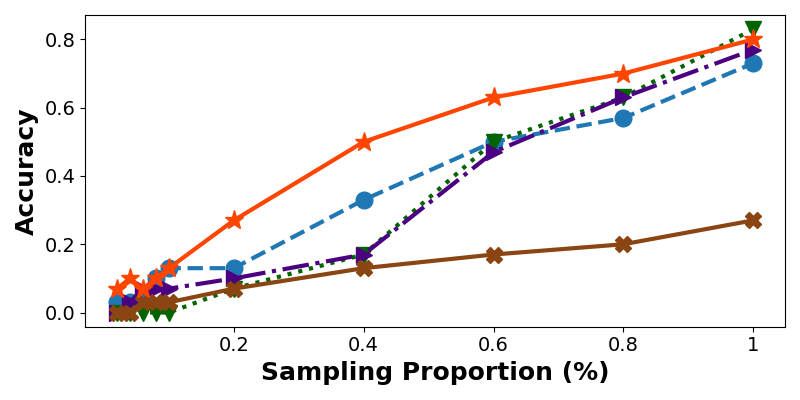}
        \caption{DB-E3 (hard)}
        \label{fig:dblp-acc-edge-3}
    \end{subfigure}
    \hfill
    \begin{subfigure}[b]{0.24\linewidth}
        \centering
        \includegraphics[width=\linewidth]{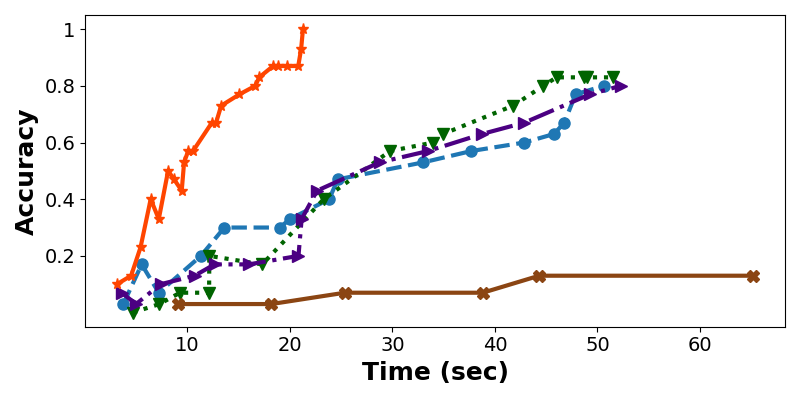}
        \caption{DB-E3 (hard)}
        \label{fig:dblp-time-edge-1}
    \end{subfigure}

    \begin{subfigure}[b]{0.24\linewidth}
        \centering
        \includegraphics[width=\linewidth]{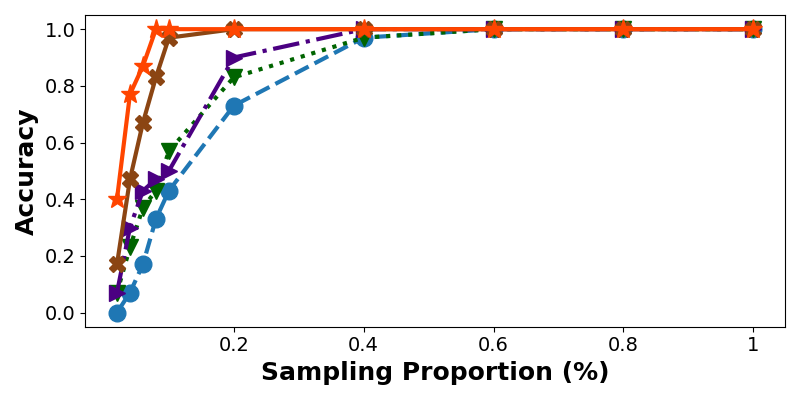}
        \caption{DB-P1 (easy)}
        \label{fig:dblp-acc-path-1}
    \end{subfigure}
    \hfill
    \begin{subfigure}[b]{0.24\linewidth}
        \centering
        \includegraphics[width=\linewidth]{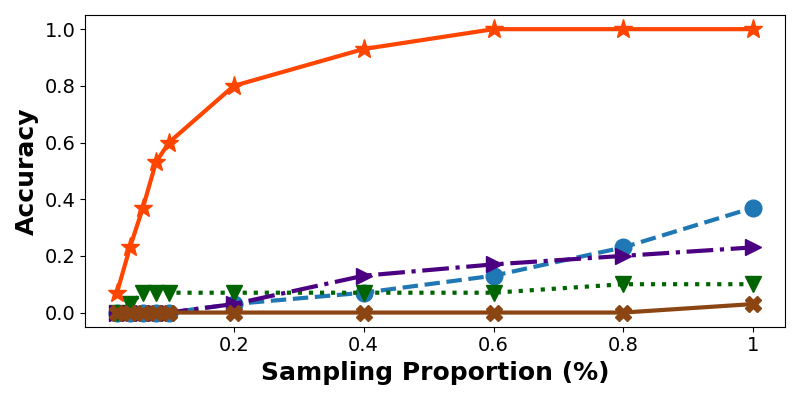}
        \caption{DB-P2 (medium)}
        \label{fig:dblp-acc-path-2}
    \end{subfigure}
    \hfill
    \begin{subfigure}[b]{0.24\linewidth}
        \centering
        \includegraphics[width=\linewidth]{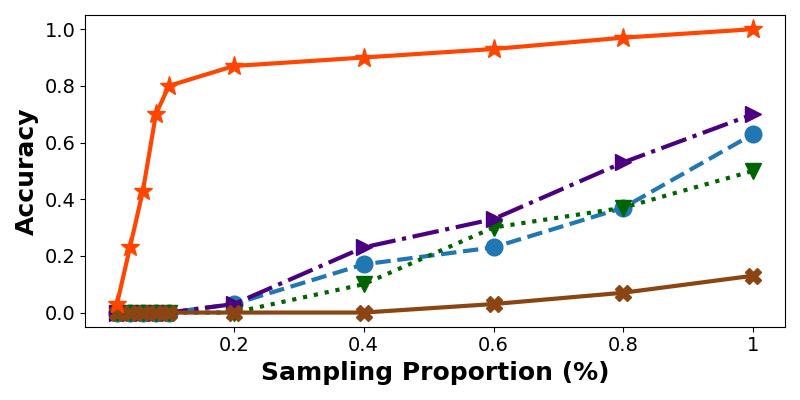}
        \caption{DB-P3 (hard)}
        \label{fig:dblp-acc-path-3}
    \end{subfigure}
    \hfill
    \begin{subfigure}[b]{0.24\linewidth}
        \centering
        \includegraphics[width=\linewidth]{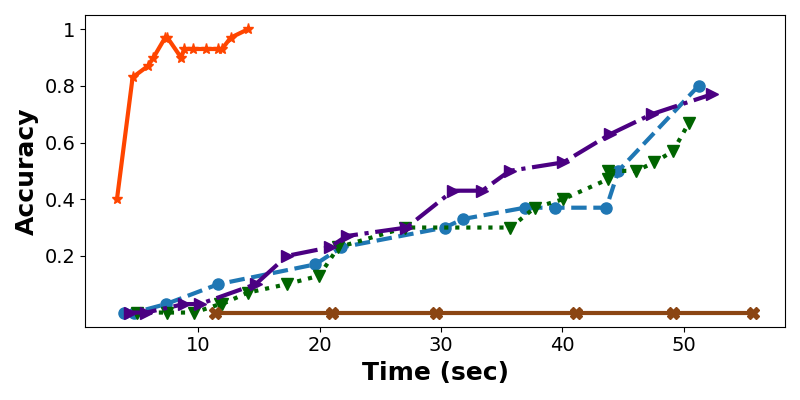}
        \caption{DB-P3 (hard)}
        \label{fig:dblp-time-path-1}
    \end{subfigure}
    \caption{Comparison of the top five sampling methods for accuracy (a-c, e-g, i-k) and time-accuracy (d, h, l) performance across three node (a-d), three edge (e-h), and three path hypotheses (i-l) for DBLP.}
    \label{fig:dblp-results}
\end{figure*}

\subsection{Accuracy}\label{expx:accuracy}
Table~\ref{tab:acc-table} presents the accuracy of 11 existing samplers 
and $\text{PHASE}_{\text{opt}}$ with a fix $B$ in column three. $B$ is set to 
ensure the accuracy is sufficiently stabilized without saturating across all samplers. Each row reports an average accuracy from three examples, based on 30 runs each. The highest and second highest accuracies are highlighted in bold and underlined, respectively.

Table~\ref{tab:acc-table} shows $\text{PHASE}_{\text{opt}}$'s robust performance across most scenarios, except for MovieLens edge hypotheses where it slightly lags behind \textit{NBRW}. \textit{SRW}, \textit{RWR}, and \textit{ShortestPathS} perform well on edge and path hypotheses, whereas \textit{RES}, \textit{RNS}, and \textit{DBS} struggle with path hypotheses. This observation aligns with their sampling mechanisms: node and edge samplers can hardly preserve the path information from $\mathcal{G}$. Also, $\text{PHASE}_{\text{opt}}$ outperforms \textit{FrontierS}, showcasing the effectiveness of the two weight functions in enhancing the sampler's hypothesis-awareness.

\begin{figure*}[t!]
    \centering
    \begin{subfigure}[b]{0.24\linewidth}
        \centering
        \includegraphics[width=\linewidth]{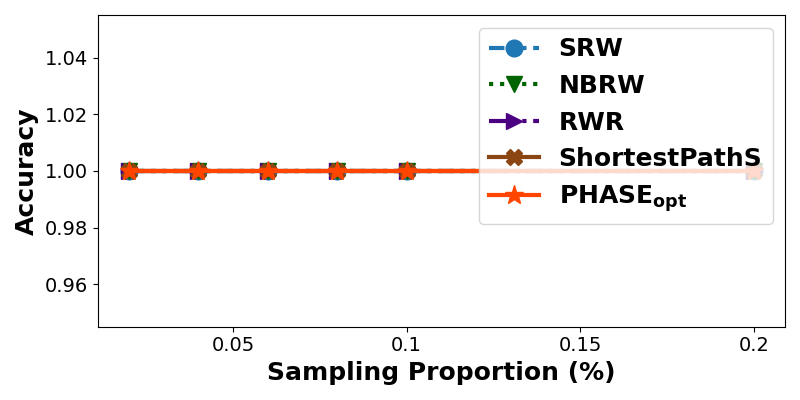}
        \caption{YP-N1 (easy)}
        \label{fig:yelp-acc-node-1}
    \end{subfigure}
    \hfill
    \begin{subfigure}[b]{0.24\linewidth}
        \centering
        \includegraphics[width=\linewidth]{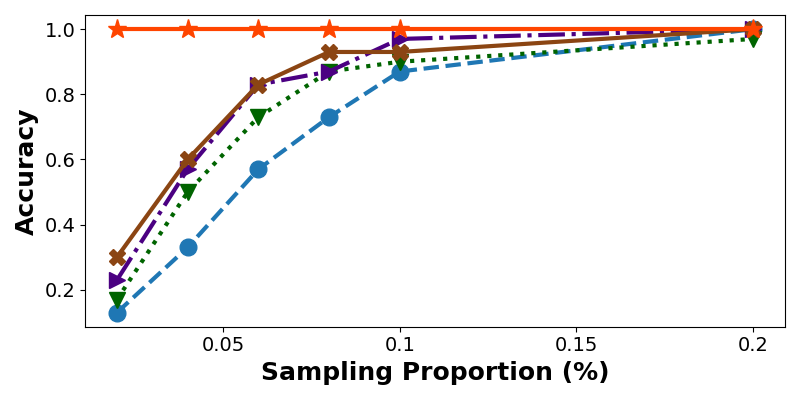}
        \caption{YP-N2 (medium)}
        \label{fig:yelp-acc-node-2}
    \end{subfigure}
    \hfill
    \begin{subfigure}[b]{0.24\linewidth}
        \centering
        \includegraphics[width=\linewidth]{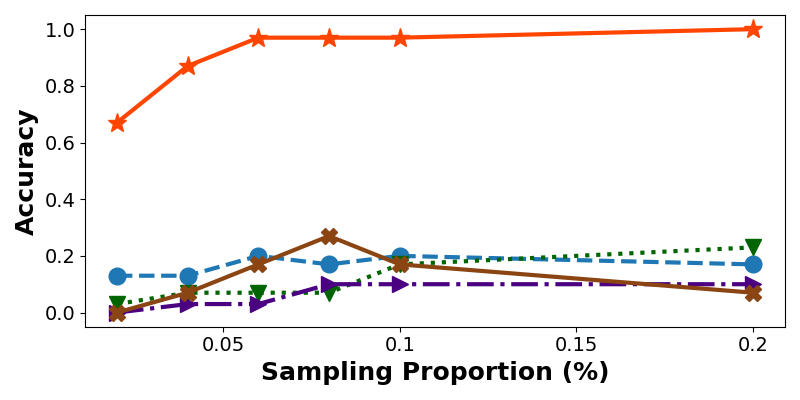}
        \caption{YP-N3 (hard)}
        \label{fig:yelp-acc-node-3}
    \end{subfigure}
    \hfill
    \begin{subfigure}[b]{0.24\linewidth}
        \centering
        \includegraphics[width=\linewidth]{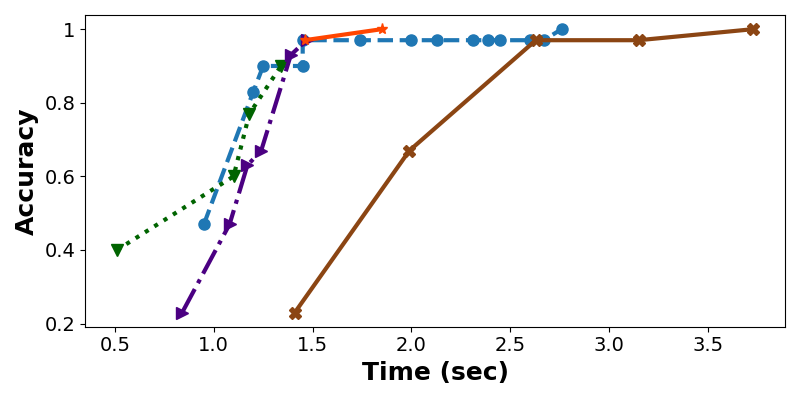}
        \caption{YP-N3 (hard)}
        \label{fig:yelp-time-node-1}
    \end{subfigure}

    \begin{subfigure}[b]{0.24\linewidth}
        \centering
        \includegraphics[width=\linewidth]{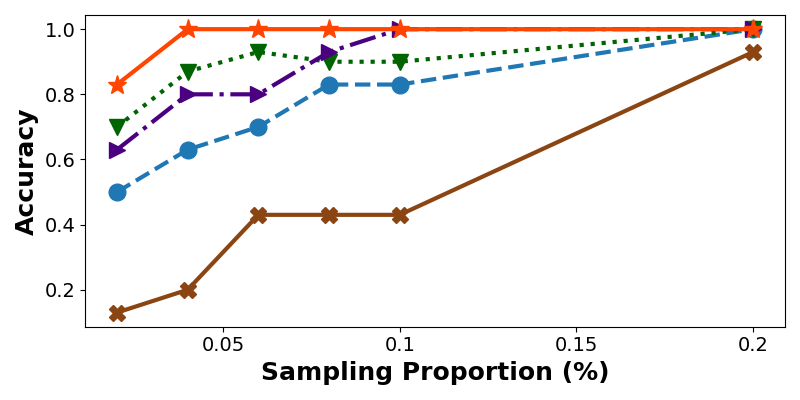}
        \caption{YP-E1 (easy)}
        \label{fig:yelp-acc-edge-1}
    \end{subfigure}
    \hfill
    \begin{subfigure}[b]{0.24\linewidth}
        \centering
        \includegraphics[width=\linewidth]{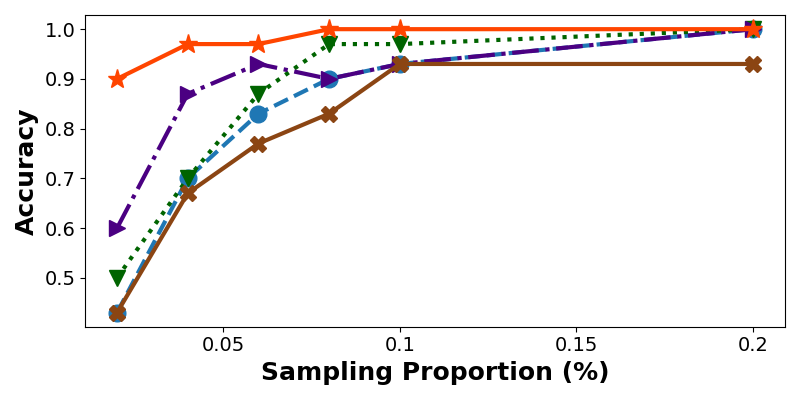}
        \caption{YP-E2 (medium)}
        \label{fig:yelp-acc-edge-2}
    \end{subfigure}
    \hfill
    \begin{subfigure}[b]{0.24\linewidth}
        \centering
        \includegraphics[width=\linewidth]{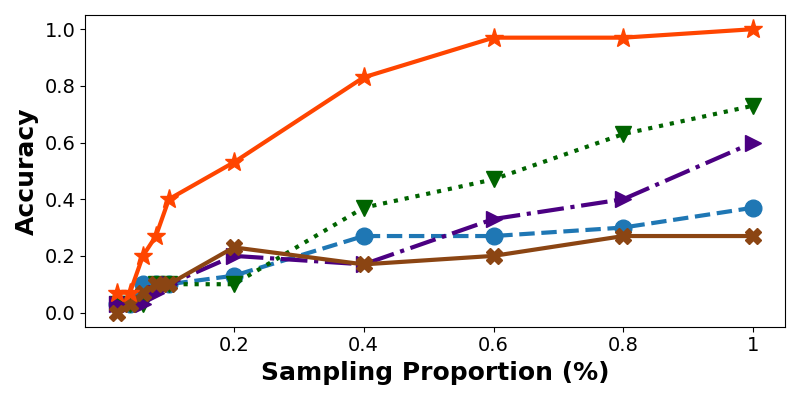}
        \caption{YP-E3 (hard)}
        \label{fig:yelp-acc-edge-3}
    \end{subfigure}
    \hfill
    \begin{subfigure}[b]{0.24\linewidth}
        \centering
        \includegraphics[width=\linewidth]{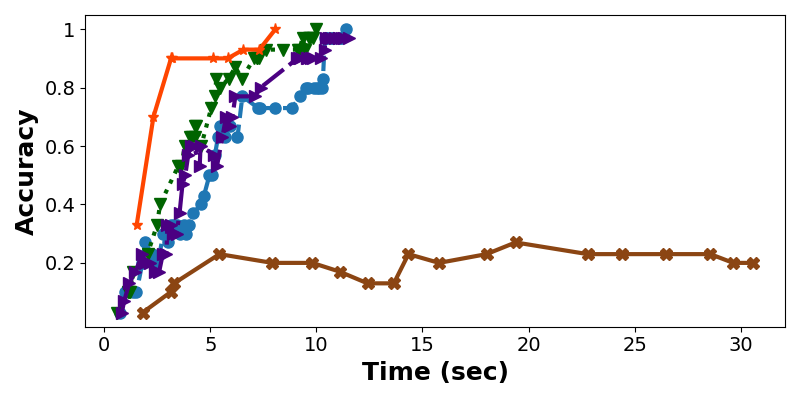}
        \caption{YP-E3 (hard)}
        \label{fig:yelp-time-edge-1}
    \end{subfigure}

    \begin{subfigure}[b]{0.24\linewidth}
        \centering
        \includegraphics[width=\linewidth]{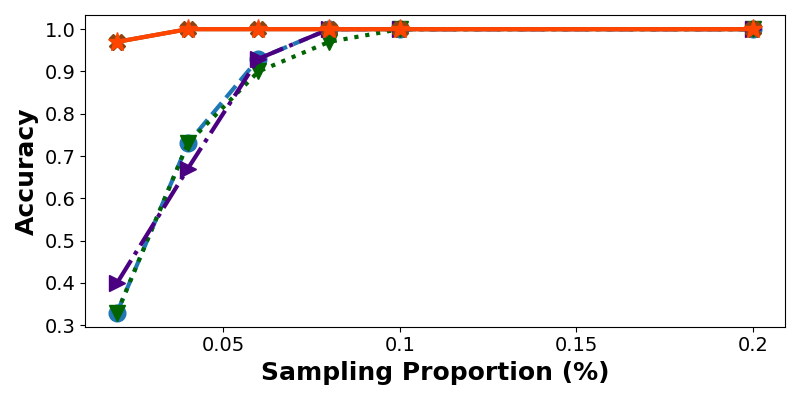}
        \caption{YP-P1 (easy)}
        \label{fig:yelp-acc-path-1}
    \end{subfigure}
    \hfill
    \begin{subfigure}[b]{0.24\linewidth}
        \centering
        \includegraphics[width=\linewidth]{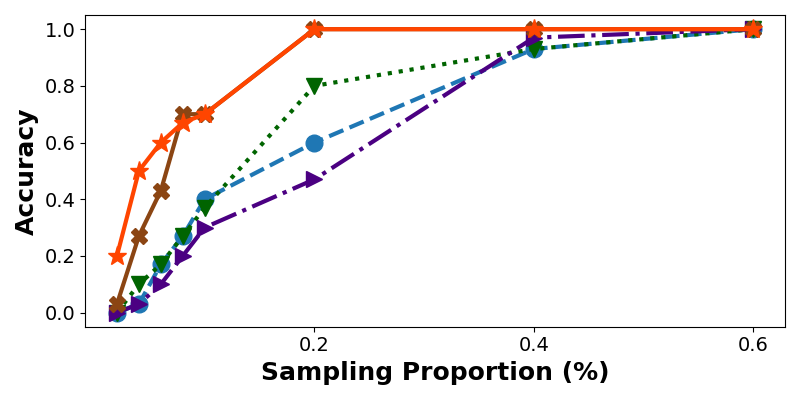}
        \caption{YP-P2 (medium)}
        \label{fig:yelp-acc-path-2}
    \end{subfigure}
    \hfill
    \begin{subfigure}[b]{0.24\linewidth}
        \centering
        \includegraphics[width=\linewidth]{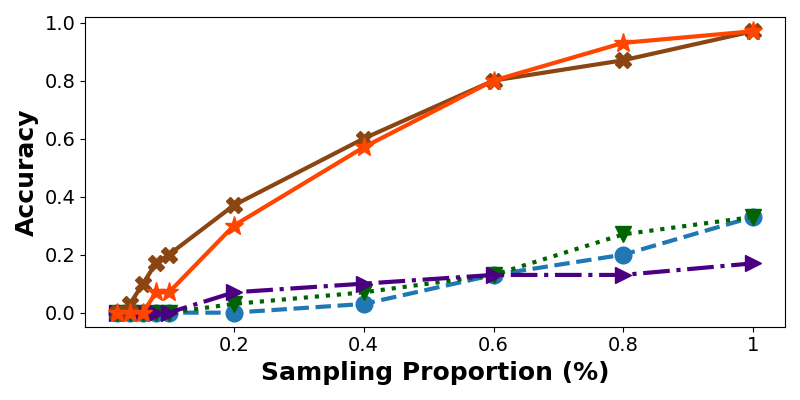}
        \caption{YP-P3 (hard)}
        \label{fig:yelp-acc-path-3}
    \end{subfigure}
    \hfill
    \begin{subfigure}[b]{0.24\linewidth}
        \centering
        \includegraphics[width=\linewidth]{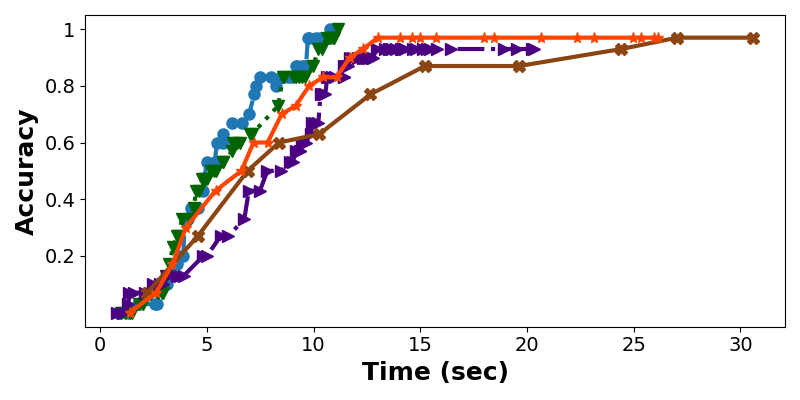}
        \caption{YP-P3 (hard)}
        \label{fig:yelp-time-path-1}
    \end{subfigure}
    \caption{Comparison of the top five sampling methods for accuracy (a-c, e-g, i-k) and time-accuracy (d, h, l) performance across three node (a-d), three edge (e-h), and three path hypotheses (i-l) for Yelp.}
    \label{fig:yelp-results}
\end{figure*}

We rank the accuracy of all samplers by averaging their accuracy per column in Table ~\ref{tab:acc-table} and identify the top five: $\text{PHASE}_{\text{opt}}$, \textit{NBRW}, \textit{ShortestPathS}, \textit{RWR}, and \textit{SRW}. The accuracy performance on individual hypotheses for DBLP and Yelp is shown in Figures ~\ref{fig:dblp-results} and ~\ref{fig:yelp-results}, respectively. Figures for MovieLens are omitted due to the space constraints and result similarity. Subfigures a-c, e-g, and i-k depict accuracy versus $B$ for three node, edge, and path hypotheses, respectively, with truncated x-axes to highlight the convergence.

In Figures~\ref{fig:dblp-results} and ~\ref{fig:yelp-results}, $\text{PHASE}_{\text{opt}}$ consistently outperforms other samplers across most hypotheses and sampling proportions. When 
there are abundant relevant nodes, edges, or paths in $\mathcal{G}$ (e.g, easy and medium hypotheses), hypothesis-agnostic samplers can perform well, reducing $\text{PHASE}_{\text{opt}}$'s relative advantage. However, its superiority is more obvious for more difficult hypotheses (e.g., hard hypotheses). Also, for difficult hypotheses with longer paths in DBLP, $\text{PHASE}_{\text{opt}}$ achieves the highest accuracy at any sampling proportion, as shown in Figures~\ref{fig:dblp-APPA-acc} and ~\ref{fig:dblp-APVPA-acc}. Moreover, \textit{ShortestPathS} sometimes shows competitive performance, possibly due to the high betweenness centrality of the concerned nodes, edges, or paths, which implies
many shortest paths traversing them.


\subsection{Scalability}\label{expx:time}
In Table ~\ref{tab:time-table}, the lowest execution time is in bold. First, the execution time increases with dataset size. Second, \textit{RNS} generally requires the shortest execution time because it uniformly samples nodes. Our method, $\text{PHASE}_{\text{opt}}$, has varying time performance across datasets. Specifically, it ranks among the highest in execution time for Yelp, especially for path hypotheses. This is attributed to the limited node types that extend path extraction times. While for DBLP, its execution time ranks in the lowest five. 

As $\text{PHASE}_{\text{opt}}$ can achieve high accuracy with a small $B$, it is unfair to compare the time efficiency at a fixed $B$. Instead, we plot accuracy versus execution time in subfigures d, h, and l of Figures ~\ref{fig:dblp-results} and ~\ref{fig:yelp-results} for DBLP and Yelp, respectively, for the top five samplers ranked by accuracy.
Due to the space constraints, only the most difficult hypothesis of each type is shown. We measure time and accuracy starting from $B = 1000$ in increments of 1000 until the accuracy reaches one, or time reaches 50s (DBLP) and 30s (Yelp). 
We find that $\text{PHASE}_{\text{opt}}$ consistently achieves high accuracy in the shortest time and with the least $B$. Also, the execution time does not grow exponentially. Moreover, when the path length increases, as shown in Figure ~\ref{fig:dblp-APPA-time} and ~\ref{fig:dblp-APVPA-time}, $\text{PHASE}_{\text{opt}}$ achieves the highest accuracy in the shortest time.

\section{Related Work}\label{sec:relwork}
There has been a lot of research interest on hypothesis testing ~\cite{hunter2008goodness, DBLP:journals/corr/abs-2212-10513,arias2014community,xia2017hypothesis,croft2011hypothesis,DBLP:conf/nips/GhoshdastidarL18,DBLP:journals/vldb/Omidvar-Tehrani19} and graph sampling techniques~\cite{DBLP:conf/kdd/LeskovecF06,DBLP:conf/sigmetrics/LeeXE12,DBLP:conf/icde/LiWL0LXL19}. In this section, we summarize the most representative works.  

\stitle{Hypothesis Testing on Graphs.} 
Hypotheses on graphs can be categorized by their object of interest~\cite{zhao2011community, arias2014community, DBLP:journals/corr/BickelS13}. Tang et al. and Ghoshdastidar et al. ~\cite{tang2017semiparametric, DBLP:conf/colt/GhoshdastidarGC17,ghoshdastidar2020two} extend the one-sample problem into two-sample to test whether two groups of random graphs are obtained by the same generative model or with the same graph distribution. In~\cite{xia2017hypothesis}, nodes are objects of interest, and one-sample hypothesis testing is used to detect conditional dependence between nodes in brain connectivity networks. 

Our goal is closely related to works that focus on nodes as the object of interest. However, there are two major differences between our work and the existing ones. First, we enable more expressive hypotheses, including edge and path hypotheses, on graphs. Second, we leverage graph sampling methods to conduct hypothesis testing.

\stitle{Sampling Methods on Graphs.} 
Most sampling methods are designed to sample a representative subgraph to accurately estimate graph properties~\cite{DBLP:conf/kdd/LeskovecF06,DBLP:journals/ton/StutzbachRDSW09,DBLP:conf/www/MaiyaB10,DBLP:conf/icdm/HublerKBG08}. Leskovec and Faloutsos ~\cite{DBLP:conf/kdd/LeskovecF06} are the first to study this problem in real-world networks~\cite{DBLP:conf/kdd/LeskovecF06} by proposing sampling techniques to maintain degree distribution, clustering coefficient, and distribution of component sizes. Hübler et al.~\cite{DBLP:conf/icdm/HublerKBG08} propose the Metropolis-Hastings sampling method. Maiya and Berger-Wolf~\cite{DBLP:conf/www/MaiyaB10} define the community representativeness sample and propose a community structure expansion sampler. Later, many sampling methods are proposed to improve the efficiency and convergence rate of simple random walk~\cite{DBLP:conf/sigmetrics/LeeXE12, DBLP:conf/icde/LiWL0LXL19}. Besides representative sampling, some sampling methods are designed for specific tasks, such as graph compression~\cite{DBLP:conf/sc/BestaWGGIOH19}, community detection~\cite{DBLP:conf/www/MaiyaB10}, and graph visualization~\cite{DBLP:conf/visualization/RafieiC05,DBLP:journals/tvcg/WuCASQC17}. However, none of them is designed for hypothesis testing on attributed graphs.
\section{Conclusion}\label{sec:conclu}

In this paper, we develop a framework for hypothesis testing on large attributed graphs, which accommodates 11 existing hypothesis-agnostic samplers and new hypothesis-aware samplers. We propose dedicated optimizations to speed up sampling while achieving high test significance and accuracy. We also demonstrate theoretically and empirically that our hypothesis-aware sampling achieves earlier convergence of the hypothesis estimator than other methods.

Our work opens several new directions in the area of hypothesis sampling on graphs. The first direction is to examine additional optimizations that make use of domain-specific information on the input graph. The second direction is to handle more expressive hypotheses and specify two-sample and multiple sample scenarios.

\begin{acks}
 Reynold Cheng, Yun Wang, and Chrysanthi Kosyfaki were supported by the University of Hong Kong (Project 109000579), the HKU Outstanding Research Student Supervisor Award 2022-23, and the HKU Faculty Exchange Award 2024 (Engineering).
\end{acks}

\bibliographystyle{ACM-Reference-Format}
\balance
\bibliography{main}

\end{document}